%% file: main.tex
\documentclass[]{fairmeta}
\usepackage[utf8]{inputenc} % allow utf-8 input
\usepackage[T1]{fontenc}    % use 8-bit T1 fonts
\usepackage{hyperref}       % hyperlinks
\usepackage{url}            % simple URL typesetting
\usepackage{microtype}      % microtypography

\usepackage{amsfonts}
\usepackage{amsmath}
\usepackage{amssymb}
\usepackage{amsthm}
\usepackage{bbold}
\usepackage{booktabs}
\usepackage{enumitem}
\usepackage{graphicx}
\usepackage{mathtools}
\usepackage{multirow}
\usepackage{nicefrac}
\usepackage{xcolor}
\usepackage{xfrac}
\usepackage{todonotes}
\usepackage{wrapfig}
\usepackage{subcaption}
\usepackage{pgfplots}
\pgfplotsset{compat=1.13}
\usepackage[most]{tcolorbox}

\input{iclr_2027/math_commands}

\newcommand{\bz}[0]{\pmb{z}}

\newcommand{\hparam}[1]{\mathtt{#1}}
\newcommand{\distribution}[1]{\operatorname{#1}}
\theoremstyle{plain}

\theoremstyle{definition}

\title{Small-Scale Experiments: Are We There Yet?}

\author[1,2]{Nicholas Lourie}
\author[2]{Kyunghyun Cho}
\author[1, \dagger]{Karen Ullrich}
\author[1, \dagger]{Sanae Lotfi}

\affiliation[1]{FAIR at MSL Meta}
\affiliation[2]{New York University}
\contribution[\dagger]{Joint Supervision}

\abstract{
    \input{contents/abstract}
}
\date{\today}
\correspondence{\\Nicholas Lourie at \email{nick.lourie@nyu.edu}, Kyunghyun Cho at \email{kyunghyun.cho@nyu.edu}, \\Karen Ullrich at \email{karenu@meta.com}, Sanae Lotfi at \email{sanaelotfi@meta.com}}

\begin{document}

\maketitle

% CONTENTS
\input{contents/introduction}
\input{contents/background}

\input{contents/estimating_scaling_laws}

\input{contents/understanding_hp_landscape}
\input{contents/designing_small_scale_experiments}
\input{contents/conclusion}

\clearpage
% AI USE STATEMENT + Acknowledgements
\input{contents/statements_and_contributions}

%  BIBLIOGRAPHY
\bibliographystyle{assets/plainnat}
\bibliography{references}

%  APPENDIX
\newpage
\beginappendix
\section*{Appendix}
\input{contents/appendix/appendix_related_work} % A
\clearpage
\input{contents/appendix/experimental_design} % B
\clearpage
\input{contents/appendix/model_of_hp_surface} % C
\clearpage
\input{contents/appendix/additional_theoretical_results} % D
\clearpage
\input{contents/appendix/additional_empirical_results} % E
\clearpage
\input{contents/appendix/case_study} % F

\end{document}

%% file: iclr_2027/math_commands.tex
\newcommand{\bX}{\mathbf{X}}
\newcommand{\bx}{\mathbf{x}}
\newcommand{\E}{\mathbb{E}}
\newcommand{\Prob}{\mathbb{P}}
\newcommand{\reals}{\mathbb{R}}
\newcommand{\normal}{\mathcal{N}}
\newcommand{\quadratic}{\mathcal{Q}}
\renewcommand{\d}{\mathrm{d}}   % upright differential 'd'; write \, before it, e.g. \int f(u)\, \d u

\usepackage{amsmath,amsfonts,bm}

\def\eqref#1{equation~\ref{#1}}
\def\1{\bm{1}}

\DeclareMathAlphabet{\mathsfit}{\encodingdefault}{\sfdefault}{m}{sl}
\SetMathAlphabet{\mathsfit}{bold}{\encodingdefault}{\sfdefault}{bx}{n}

%% file: contents/abstract.tex
Scaling laws promised cost-effective experiments; six years later, they have yet to fully deliver. Instead, researchers have found them unreliable at small scales, starting at 4M parameters, and concluded that sizable models cannot be avoided. We show this is not the case and that the confounding factor is hyperparameter tuning. In fact, small models are highly sensitive, but hyperparameter sensitivity fades with scale. This small-scale sensitivity makes scaling laws easy to miss because they only emerge on the fully tuned frontier, and reaching that frontier requires an extensive search far beyond what most ever run. By ablating the basic scaling law recipe, we show well-tuned hyperparameters matter more than any other ingredient. Further, we show that as scale increases, the hyperparameter loss surface becomes lower dimensional, making it easier to find those hyperparameters. Nevertheless while scaling laws exist in small models, extrapolation hits statistical limitations and a holistic approach becomes necessary. We apply our insights to a question that once took the field years to settle: where to place normalization layers in the transformer architecture. Using a rigorous experimental setup inspired by our findings, we find that pre-normalization works better as models grow in size. With the right tools and a better understanding, small-scale experiments can deliver on scaling laws' long-awaited promise.

%% file: contents/introduction.tex
\section{Introduction}
\label{sec:introduction}

Foundation models have revolutionized the way we think about computing. They arose as a paradigm only five years ago \citep{Bommasani2021FoundationModels}, yet in that time have become the basis for new discoveries across a broad set of scientific fields \citep{Jumper2021HighlyAccurate, Lam2023LearningSkillful, Merchant2023ScalingDeep, Trinh2024SolvingOlympiad, Hubert2026Olympiad}. As you might expect, those five years saw a flurry of research into how best to build these models, and one key ingredient emerged: scale. Unfortunately, scale makes experiments incredibly expensive.

The natural solution is to experiment at the small scale, then transfer to the large scales. Despite the early promise of scaling laws \citep{Kaplan2020ScalingLaws}, this solution has not yet materialized after half a decade. Instead, pretraining research grew exponentially in cost, with experiments (not the final training run) dominating the overall expense of model development \citep{cottier2025risingcoststrainingfrontier, epoch2026randdvstrainingcompute}.

Scaling laws failed to deliver because they hit a major roadblock, which is that they are notoriously elusive at small scales. Famously, the two most influential works in the literature reached contradictory conclusions. \citet{Kaplan2020ScalingLaws} held parameters should scale much faster than data, as a result early language models were under-trained \citep{Brown2020Language, rae2022scalinglanguagemodelsmethods, Chowdhery2023PaLM}; \citet{Hoffmann2022AnEmpirical} found parameters and data should scale at the same rate. \citet{Porian2024Resolving} traced this discrepancy to methodological choices such as whether to count embeddings in the parameters, how long to warm up the learning rate, and scale-dependent hyperparameter tuning; choices that disproportionately affect small-scale models \citep{pearce2024reconciling}. This early discord foreshadowed later difficulty. Surveying the literature, \citet{li2025misfitting} found many scaling laws hard to reproduce and, in particular, that scaling laws based on smaller models with 100M parameters or less are unreliable. \citet{Li2025PredictableScale} aptly summarize the situation as the ``\emph{scaling gap} where insights from small-scale experiments often fail to transfer to resource-intensive production systems''.

How do we close this gap and deliver on scaling laws' original promise? We find that the recipe for a scaling law has one key ingredient that, more than any other, determines final quality: rigorous hyperparameter tuning (\S\ref{sec:estimating-small-scale-scaling-laws}). By ablating different choices, we show scaling laws are fairly robust to details like how to count parameters or whether to decay the learning rate (\S\ref{sec:estimating-small-scale-scaling-laws:the-impact-of-methodology}). However, poorly tuned hyperparameters can be disastrous (\S\ref{sec:estimating-small-scale-scaling-laws:the-importance-of-hyperparameter-tuning}). Whereas common practice tunes over a small grid, you may need to search hundreds of configurations to reveal the scaling law in small models. Such a search sounds expensive, yet we also find scaling laws extend remarkably far down; as little as 4M parameters where models train in under an hour on a single GPU, bringing the cost well within reach for many academic and industry labs.

Hyperparameters unduly affect small-scale laws because small-scale models are unduly difficult to tune. We explain this difficulty in terms of the hyperparameter loss surface (\S\ref{sec:understanding-the-hyperparameter-loss-surface-across-scales}). As parameters and data grow together, they drive good configurations to fill a larger fraction of the space and the intrinsic dimension of the surface goes down (\S\ref{sec:understanding-the-hyperparameter-loss-surface-across-scales:hyperparameter-tuning-gets-easier-with-scale}---\S\ref{sec:understanding-the-hyperparameter-loss-surface-across-scales:the-hyperparameter-loss-surface-becomes-lower-dimensional}). Moreveover, as model size increases, the \emph{effective} number of hyperparameters drops to one. This phenomenon introduces a new way of thinking about scaling, where one should do a thorough job sweeping hyperparameters in the smaller scales to due to their sensitivity whereas larger models are easier to tune (\S\ref{sec:understanding-the-hyperparameter-loss-surface-across-scales:implications-for-small-scale-experiments}). Luckily smaller scale experiments are significantly cheaper to run than their larger scale counterparts so proper scaling becomes a matter of proper computational budget allocation. 

Building on our insights, we propose a new methodology for proper scaling drawing on three tools from the literature (\S\ref{sec:designing-small-scale-experiments}). Namely, the \emph{noisy quadratic limit} reveals if hyperparameters are fully tuned and how hard that is to accomplish \citep{lourie2025hyperparameter}; \emph{scaling laws} describe how loss changes as models scale up \citep{rosenfeld2020a, Kaplan2020ScalingLaws, Hoffmann2022AnEmpirical}; and the \emph{perplexity--capability correspondence} answers whether conclusions about pretraining loss have consequences for the capabilities we ultimately care about \citep{mayilvahanan25a}.\footnote{
    Even when we often can not predict these capabilities in advance \citep{lourie-etal-2025-scaling}.
} Small-scale experiments exploit strong assumptions and we argue for an approach where each assumption becomes a diagnostic that improves our qualitative understanding of how the model scales (\ref{sec:designing-small-scale-experiments:a-methodology-for-small-scale-experiments}). This qualitative understanding is important because even though phenomena like scaling laws exist at the small scale, quantitative techniques such as extrapolating the loss compound small errors (\ref{sec:designing-small-scale-experiments:the-pitfalls-of-pure-extrapolation}). Finally, we apply our takeaways to a question that once took the field years to settle (\ref{sec:designing-small-scale-experiments:case-study-pre-norm-vs-post-norm-architectures}): where normalization layers should go in the transformer architecture, and find that larger scales favor pre-normalization \citep{Vaswani2017Attention, baevski2018adaptiveinputrepresentationsneural, Xiong2020OnLayer}.

In the end, these ideas give a practical methodology for model-centric research, which  improves the model holding \textit{data fixed} in contrast to \emph{data-centric} research, which improves the data holding the model fixed. 
The methodology is as follows: first build a complete understanding at the small scale by thoroughly exploring the model's hyperparameters, then use simple rules to carry it up. This regimen offers the rare opportunity to cut costs and raise rigor, but it takes the right tools discussed in this work. With them, small-scale experiments can finally deliver on their long-awaited promise.
We note finally that our discussion focuses on model-centric research while data-centric research likely requires different techniques (\S\ref{sec:conclusion}). 

%% file: contents/background.tex
\section{Background and related work}
\label{sec:background}

\subsection{Scaling Laws}
\label{sec:background:scaling-laws}

Researchers have long sought to predict the outcomes of large-scale training from small-scale experiments. Early work studied how models improve with more data \citep{cortes1993learning, hestness2017deep, sun2017revisiting}; later work, with the arrival of web-scale data, turned towards parameters and compute as scaling laws became the primary tool for managing the rising cost of experiments \citep{rosenfeld2020a, Kaplan2020ScalingLaws}. After \citet{Kaplan2020ScalingLaws} demonstrated that pretraining loss falls predictably over many orders of magnitude for language models, similar scaling laws were soon discovered in many other domains \citep{henighan2020scalinglawsautoregressivegenerative, jones2021scalingscalinglawsboard, Cherti_2023_CVPR}. Such scaling laws offered an enticing promise: study a model cheaply at the small scale, then extrapolate to the large.
 
Classically, scaling laws  relate parameters ($p$) and data ($d$) to pretraining loss ($\mathcal{L}$). \citet{Hoffmann2022AnEmpirical} popularized the most widely used functional form \citep{rosenfeld2020a}:
\begin{equation}\label{eq:scaling-law_joint}
    \mathcal{L}(p, d) = \epsilon + \frac{\zeta}{p^\iota} + \frac{\eta}{d^\kappa}
\end{equation}

During training the network runs on every example, so compute is proportional to parameters times data: $c \propto p d$. With  \Cref{eq:scaling-law_joint}, this constraint gives the compute-optimal loss as a power law: $\mathcal{L}(p, d) - \epsilon \propto c^{-\frac{\iota\kappa}{\iota + \kappa}}$ given a fixed compute budget, see \Cref{app:scaling-laws} for details. 

However, the promise of predictable large scale performance for fixed compute budget proved difficult to keep. \citet{pearce2024reconciling} found small-scale experiments led to discrepancies across major scaling law studies \citep{Kaplan2020ScalingLaws, Hoffmann2022AnEmpirical}. More recently, \citet{li2025misfitting} surveyed the literature, found scaling law estimates hard to reproduce, and reported those built from small models are unreliable. We revisit this regime and reach a more optimistic conclusion: with rigorous tuning, scaling laws exist even at a few million parameters though we must extrapolate them with care.

\subsection{Hyperparameter Optimality}
\label{sec:background:the-noisy-quadratic-limit}

With the evidence presented in the previous section, we might develop the intuition that tuning hyperparamerters is one of the most important problems in estimating accurate scaling laws. 
If that is true, how do we know if hyperparameters are fully tuned? 
\citet{lourie2025hyperparameter} found that near the optima, hyperparameter-loss surfaces can be well approximated with a simple low rank model; the noisy quadratic.
Empirically, this approximation holds for a large region around the optimum. 
Since this region determines the practical questions we care about, like tuning difficulty or the best attainable performance, we will use the noisy quadratic throughout this study. For more details see \Cref{app:background:the-noisy-quadratic-limit}.

After drawing conclusions at the small scale, we still need the right hyperparameters to scale up. One strategy parametrizes them so that their optimal values stay \textit{fixed} across scales  \citep{yang2021tuning, yang2024tensor, dey2025completep, mlodozeniec2026completed}. A complementary approach fits scaling laws for the optimal values themselves \citep{deepseekai2024deepseekllmscalingopensource, li2025predictablescaleistep, Bergsma2025PowerLines}. 
Many methods work reasonably well. We will later see that is is likely because as models grow, they become less sensitive to their hyperparameters thus a configuration tuned cheaply at the small scale transfers well with just a little adjustment --- a central empirical finding in this study we will detail later.

%We will try to model the hyperparameter progression through scales with the noisy quadratic, similar in spirit to \citet{zhang2026configurationtoperformancescalinglawneural} who predict performance from scale together with the full hyperparameter configuration but different in that they fit across scales using publicly available training runs, an approach better suited to modeling familiar configurations than evaluating wholly new ideas.

\subsection{Perplexity--Capability Correspondence}
\label{sec:background:perplexity-capability-correspondence}

Scaling laws' successes in pretraining inspired many to seek them for downstream tasks \citep{ivgi-etal-2022-scaling, gadre2025language, chen2025scaling,Mahmood_2022_CVPR, openai2023gpt4}. This stream of work delivered mixed results however as downstream laws inherit and compound the tuning difficulties of upstream ones, spread across multiple training stages and a battery of evaluations. As a consequence, downstream tasks exhibit a wide range of unpredictable behavior, such as emergent \citep{wei2022emergent,srivastava2023beyond}, inverse \citep{mckenzie2023inverse, wilcox2024bigger}, or even U-shaped scaling \citep{wei2023inversescalingushaped}. \citet{lourie-etal-2025-scaling} found as little as 39\% of tasks scale reliably. Carefully chosen metrics can mitigate some of this irregularity \citep{schaeffer2023mirage, schaeffer2025why} but not eliminate it \citep{du-etal-2024-understanding, zhao2026randomscalingemergentcapabilities}. 

To sidestep this issue, researchers recast downstream performance as a function of pretraining loss instead of data and compute \citep{xia-etal-2023-training, snell2024predicting, huang2024compression, du-etal-2024-understanding, gadre2025language, chen2025scaling}. Remarkably, \citet{mayilvahanan25a} showed pretraining loss corresponds to capability as long as the pretraining data is held fixed across changes in scale, hyperparameters, architecture, and even the tokenizer. In other words: models of the same perplexity obtain the same capabilities, even when those capabilities are hard to know in advance. We term this phenomenon \emph{perplexity--capability correspondence}. Hence in this study we will rely on and assume this correspondence; a practical illustration is in Appendix~\ref{sec:appendix-Downstream-Scaling-laws}.

%% file: contents/estimating_scaling_laws.tex
\section{Estimating (Small-Scale) Scaling Laws}
\label{sec:estimating-small-scale-scaling-laws}

In this study, we will use models ranging in size from 4--34M to estimate scaling laws, 67M--134M for validation, and 268M for testing.
All details about the experimental design chosen in this study are in Appendix~\ref{app:experimental-design}.

What does it take to estimate scaling laws from small-scale experiments? We find them fairly robust to different decisions (\S\ref{sec:estimating-small-scale-scaling-laws:the-impact-of-methodology}), but one ingredient makes or breaks the resulting law: hyperparameter tuning (\S\ref{sec:estimating-small-scale-scaling-laws:the-importance-of-hyperparameter-tuning}).

\subsection{The Impact of Methodology}
\label{sec:estimating-small-scale-scaling-laws:the-impact-of-methodology}

Despite the frequent debate over parameter definitions, we find they make little difference. Beyond that, we study three refinements on a basic approach: tuning hyperparameters per token budget, decaying the learning rate, and tying the scaling exponents. Some help but none, we will see, is what makes the estimate work.

\begin{figure}[t!]
    \centering
    \includegraphics[width=\linewidth]{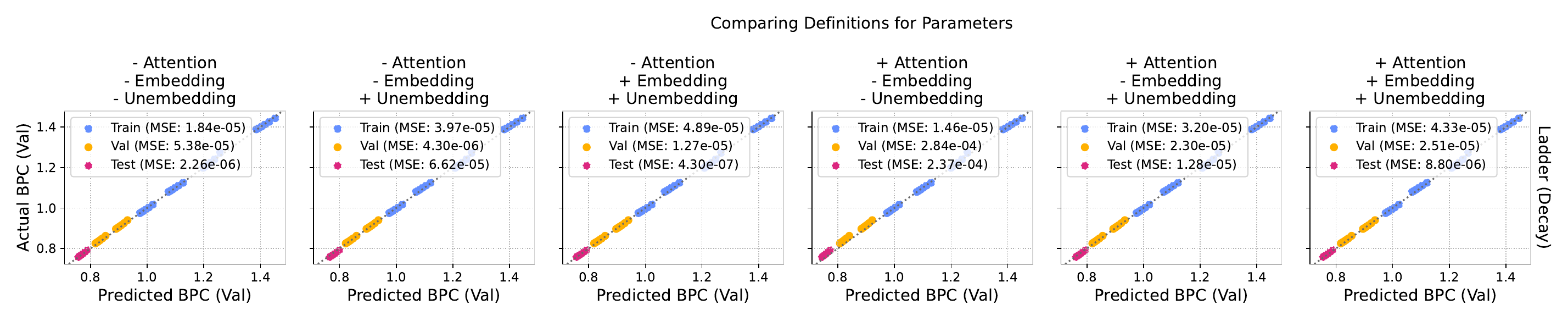}
    \caption{
        \emph{Even at small scales, how you count parameters has limited impact on the scaling law relative to other factors.} Researchers count parameters in different ways, diverging on whether to (1) account for attention's FLOPs (\texttt{Attention}), (2) include the embedding (\texttt{Embedding}), and (3) include the unembedding (\texttt{Unembedding}). The embedding and unembedding have the same number of parameters, giving six distinct counts. Each one yields a reasonable scaling law.
    }
    \label{fig:comparing-definitions-for-parameters}
\end{figure}

\paragraph{Parameter definitions only make a small difference.} When it comes to scaling laws, researchers define parameters many different ways \citep{Kaplan2020ScalingLaws, Hoffmann2022AnEmpirical, deepseekai2024deepseekllmscalingopensource}. These definitions differ along three axes: whether to account for attention, the embedding, and the unembedding \citep[\S B]{Porian2024Resolving}. The disagreement pivots on the importance of parameters vs. compute: attention uses more FLOPs on longer contexts, while the embedding layer uses no FLOPs at all. \Cref{fig:comparing-definitions-for-parameters} reveals these differences in philosophy lead to little difference in outcome. We use the effective parameters (\texttt{Attention} and \texttt{Unembedding}, no \texttt{Embedding}) as it corresponds most directly to FLOPs per token, keeping $c = 6pd$ exact.

\begin{figure}[b!]
    \centering
    \includegraphics[width=\linewidth]{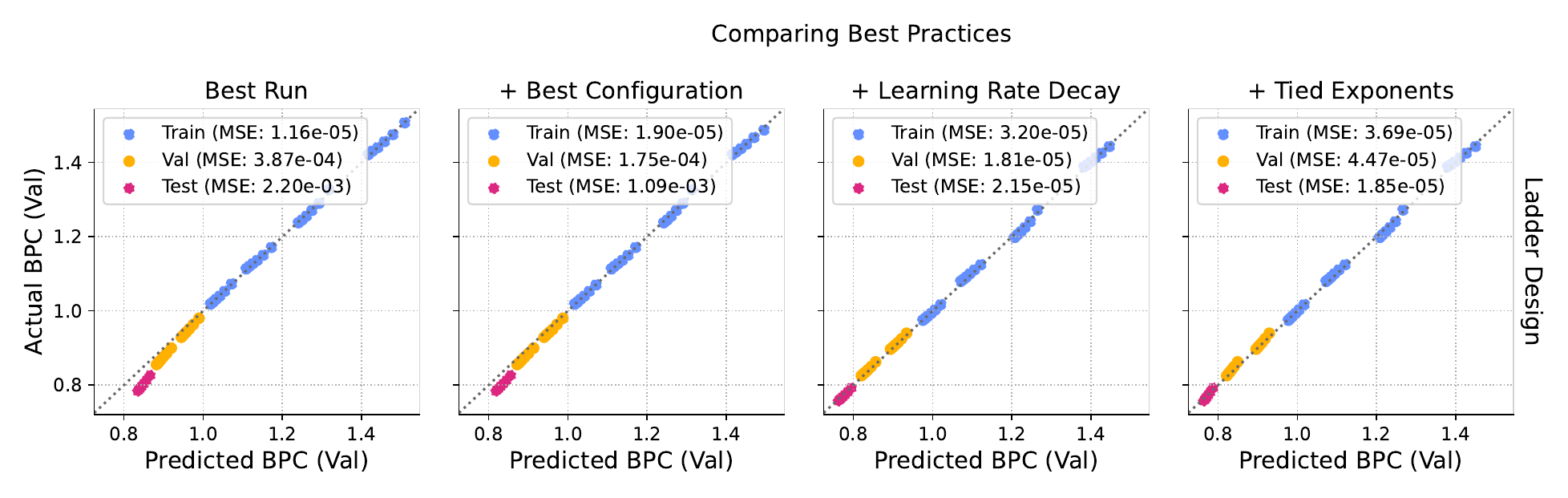}
    \caption{
        \emph{Better methodology improves the scaling law, but does not make or break it.} A recognizable law emerges from the outset and extrapolation steadily improves as we move from fitting the best run at each scale (left), to the best configuration for each parameter--data budget (middle left), to decaying the learning rate (middle right). Unlike the other refinements, tying the exponents (right) gives inconsistent results: it slightly improves test but hurts validation.
    }
    \label{fig:comparing-best-practices}
\end{figure}

\paragraph{Other choices refine the estimate, rather than enable it.} \Cref{fig:comparing-best-practices} compares increasing refinements on a basic approach. Fitting the best run at each scale without learning rate decay, the scaling law is immediately recognizable (if imprecise). Improving the methodology refines its precision: tuning hyperparameters for each parameter--data budget reduces test MSE by 50\%, while decaying the learning rate reduces it by a dramatic 98\%. In contrast, tying the exponents ($\iota = \kappa$) makes little difference. The improvements are important for obtaining a good law from small-scale experiments, but even without them the law is present.

\subsection{The Importance of Hyperparameter Tuning}
\label{sec:estimating-small-scale-scaling-laws:the-importance-of-hyperparameter-tuning}

\begin{figure}[t!]
    \centering
    \includegraphics[width=\linewidth]{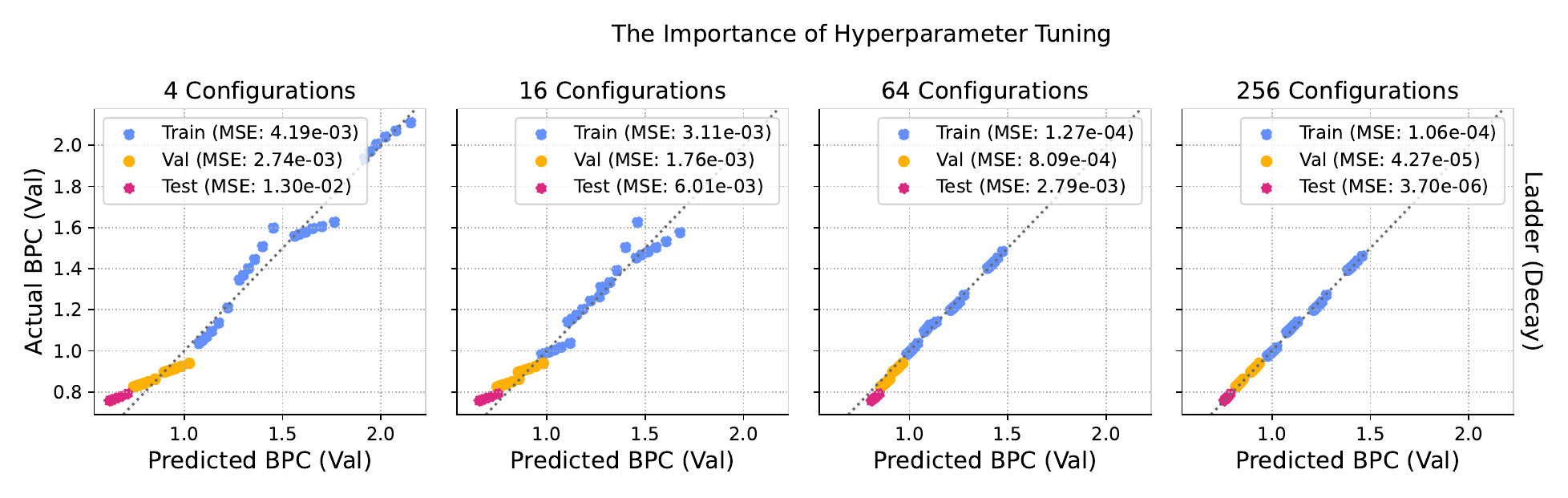}
    \caption{
        \emph{Hyperparameter tuning is the single most important step in estimating a scaling law.} Each panel fits a scaling law with a more substantial sweep; all panels evaluate (\texttt{val} and \texttt{test}) on the best configurations from our entire sweep. With 4 (left) or 16 configurations (middle left), the law is not even visible. Increasing to 64 (middle right), it clearly emerges but is imprecise. Only at 256 configurations (right) do we obtain an accurate law.
    }
    \label{fig:the-importance-of-hyperparameter-tuning}
\end{figure}

\Cref{fig:the-importance-of-hyperparameter-tuning} shows what happens with too little tuning effort. With 4 configurations per scale, the scaling law is completely absent. With 16, the scaling law has yet to fully appear; it cannot be estimated reliably as the suboptimal configurations inject too much noise. At 64, the law is clearly visible though extrapolation is weak. Only at 256 configurations do we obtain an accurate scaling law. In our previous study, each refinement improved precision on an already recognizable law; here, with too few configurations there is nothing to fit.

Thus one ingredient above all is required to estimate a scaling law, rigorous hyperparameter tuning. This fact explains why small-scale laws are so easy to miss: uncovering them demands a search far more extensive than most ever run. But why are small models so much more sensitive to their hyperparameters? We will show this change in sensitivity comes from how the hyperparameter loss surface evolves as models scale.

%% file: contents/understanding_hp_landscape.tex
\section{Understanding the Hyperparameter Loss Surface Across Scales}
\label{sec:understanding-the-hyperparameter-loss-surface-across-scales}

Throughout this section, we will make use of the same experiments we ran in the previous section. Note, that we additionally fit a noisy quadratic distribution to each scale for our analysis. 
We show that models become less sensitive to hyperparameters as they scale (\S\ref{sec:understanding-the-hyperparameter-loss-surface-across-scales:hyperparameter-tuning-gets-easier-with-scale}). This phenomenon traces back to the hyperparameter loss surface: larger scale makes it lower dimensional (\S\ref{sec:understanding-the-hyperparameter-loss-surface-across-scales:the-hyperparameter-loss-surface-becomes-lower-dimensional}). These facts influence how we design experiments (\S\ref{sec:understanding-the-hyperparameter-loss-surface-across-scales:implications-for-small-scale-experiments}): small scales require extensive search, while large scales are easy to adapt.

\subsection{Hyperparameter Tuning Gets Easier with Scale}
\label{sec:understanding-the-hyperparameter-loss-surface-across-scales:hyperparameter-tuning-gets-easier-with-scale}

\begin{figure}
    \centering
    \begin{minipage}{0.24\linewidth}
        \centering
        \includegraphics[width=\linewidth]{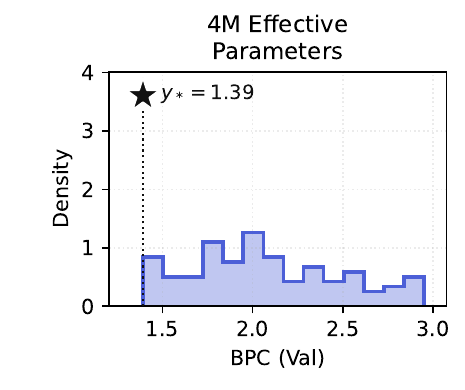}\\
        {\small (a) Scores at 4M}
    \end{minipage}
    \hfill
    \begin{minipage}{0.24\linewidth}
        \centering
        \includegraphics[width=\linewidth]{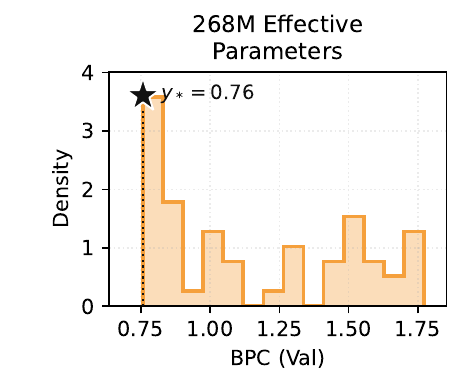}\\
        {\small (b) Scores at 268M}
    \end{minipage}
    \hfill
    \begin{minipage}{0.24\linewidth}
        \centering
        \includegraphics[width=\linewidth]{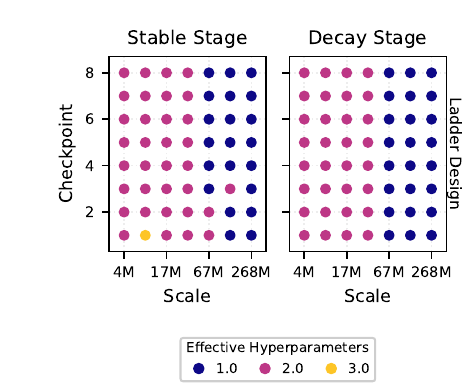}\\
        {\small (c) Effective hyperparameters}
    \end{minipage}
    \hfill
    \begin{minipage}{0.24\linewidth}
        \centering
        \includegraphics[width=\linewidth]{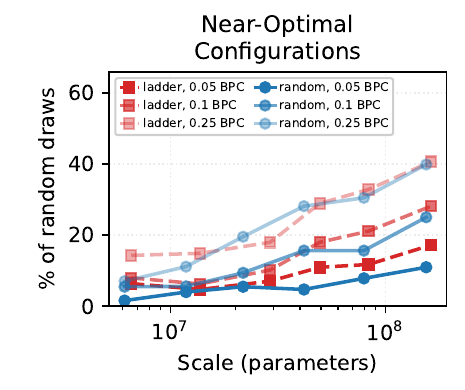}\\
        {\small (d) Near-optimal configurations}
    \end{minipage}
    \caption{
        \emph{Smaller scales are more sensitive to hyperparameter tuning.}
        We visualize the distribution over validation losses at the end of training from random
        search, in bits per character. \textbf{(a)} At 4M the
        distribution is broad and flat, so fewer configurations land near the best achievable loss,
        $y_*$.  \textbf{(b)} At 268M, the density concentrates into a peak close to $y_*$, three times higher
        on the shared axis than the 4M scale. These results demonstrate that more hyperparameter configurations land next to the optimum at larger scales and that smaller scales are more sensitive to hyperparameter tuning. \textbf{(c)} The effective hyperparameter count ($\gamma$) is the intrinsic dimension of the hyperparameter loss surface. We estimate $\gamma$ at each parameter--data budget by fitting the noisy quadratic distribution to results from random search. With and without learning rate decay, $\gamma$ trends lower as parameters increase, while increasing data has much less of an effect.
        \textbf{(d)} The percentage of configurations within $0.1$ bits per character of $y_*$ grows from 7.9\% to 28.1\%
        between 4M and 134M.
    }
    \label{fig:hyperparameter-sensitivity-loss-surface}
\end{figure}

With scale, good configurations become more common; thus, larger models trained longer are easier to tune.

\paragraph{Near-optimal configurations fill a larger fraction of the space.} The effort required to find a good configuration depends on the fraction of the space they take up. We can measure this volume via Monte Carlo integration: sample configurations uniformly and check what percent land near the optimum. This procedure is exactly random search. \Cref{fig:hyperparameter-sensitivity-loss-surface}(a,b) shows how the score distribution changes with scale: as models grow, the distribution concentrates on the best achievable loss and the density forms a peak. Thus scale makes models easier to tune.

\paragraph{Hyperparameter sensitivity decreases through both parameters and data.} We show in \Cref{fig:hyperparameter-sensitivity-loss-surface}(d) that the percentage of configurations within different loss thresholds of the optimal loss increases with scale for both the ladder and random designs.
Additionally, neither parameters nor data alone is responsible for this decrease in sensitivity; the effect comes from both. In fact, \Cref{fig:hyperparameter-sensitivity-decreases-through-both-parameters-and-data} shows the minimum, 10th, and 25th percentile validation losses from random search and illustrates how sensitivity decreases across each. The gap from the 25th percentile to the minimum directly measures sensitivity because a small gap means that a quarter of all draws are already close to optimal. Scaling one dimension while the other is held small lowers the loss and slightly moves this gap, but scaling one while the other is large collapses it. 

\subsection{The Effective Number of Hyperparameters Decreases with Scale}
\label{sec:understanding-the-hyperparameter-loss-surface-across-scales:the-hyperparameter-loss-surface-becomes-lower-dimensional}

Intuitively, a model with more hyperparameters is harder to tune; in practice, models can have many while their effective number is still low. Formally, the effective number of hyperparameters ($\gamma$) is the surface's intrinsic dimension at the optimum, a major determinant of tuning difficulty \citep{lourie2025hyperparameter}. We estimate $\gamma$ at each parameter--data budget by fitting a noisy quadratic to the tail from random search (all losses below a threshold, $\theta$). We show these estimates against model size in \Cref{fig:hyperparameter-sensitivity-loss-surface}(d), where \textbf{larger models have fewer effective hyperparameters}, dropping to one. Moreover, checkpoints across training largely share the same intrinsic dimension, while increasing parameters clearly reduces $\gamma$. 

% Therefore, parameters more than data drive the decrease in effective hyperparameters.

Thus the effective number of hyperparameters shrinks as parameters grow. Data seems to have less effect. Interestingly, prior work found the effective hyperparameters remains constant when using the same search space across different architectures \citep[\S D]{lourie2025hyperparameter}. It seems counter-intuitive that $\gamma$ depends on scale but not architecture. Moreover, scale's effect might change under an alternative parametrization, such as maximum update \citep{yang2021tuning}. All these factors likely influence $\gamma$ at some point, our results refine a prior for typical regimes. Better understanding these limits would be a valuable direction for future work.

\subsection{Implications for Small-Scale Experiments}
\label{sec:understanding-the-hyperparameter-loss-surface-across-scales:implications-for-small-scale-experiments}

The HP loss surface is not a black box; it has rich qualitative structure. We showed empirically that as parameters and data grow, it gets easier to optimize, good configurations fill the space, and the intrinsic dimension goes down. Good reason to hope that there is a simple structure underpinning the HP loss surface, which we may exploit when optimizing at large scales. After all, \citep{lourie2025hyperparameter} already showed the noisy quadratic to be a good model for the outcomes of HP optimization at a single scale. Our efforts to model the HP surface across scales however showed no consistent success. We describe several interpretable models in \S\ref{app:interpretable-models-of-the-hyperparameter-loss-surface} (and tried many more). While simple models capture coarse features of the surface, none characterize its finer details. The simplest model, one ignoring how HPs change with scale, is already a strong baseline. This outcome agrees with the qualitative picture: extrapolating to larger scales, HPs have less impact and little variation is left to model.
As a result, we treat HP optimization independently per scale.

%% file: contents/designing_small_scale_experiments.tex
\section{Designing Small-Scale Experiments}
\label{sec:designing-small-scale-experiments}

We synthesize our discussion into a methodology for small-scale experiments (\S\ref{sec:designing-small-scale-experiments:a-methodology-for-small-scale-experiments}). It emphasizes a holistic view because while regularities exist across scales, extrapolation hits statistical limits (\S\ref{sec:designing-small-scale-experiments:the-pitfalls-of-pure-extrapolation}). We demonstrate the approach on a case study from the literature: comparing pre-norm and post-norm transformers (\S\ref{sec:designing-small-scale-experiments:case-study-pre-norm-vs-post-norm-architectures}).

\subsection{A Methodology for Small-Scale Experiments}
\label{sec:designing-small-scale-experiments:a-methodology-for-small-scale-experiments}

Our approach relies on three facts:
\begin{enumerate}[noitemsep]
    \item with the pretraining data held fixed, capabilities depend on pretraining loss alone (\S\ref{sec:background:perplexity-capability-correspondence}),
    \item with rigorous hyperparameter tuning, scaling laws for pretraining extend to tiny scales (\S\ref{sec:estimating-small-scale-scaling-laws}),
    \item and as models scale, they grow far less sensitive to their hyperparameters (\S\ref{sec:understanding-the-hyperparameter-loss-surface-across-scales}).
\end{enumerate}
These principles prescribe a methodology for model-centric research: tune and measure at the small scale, understand how pretraining loss changes, then carry the winner up. The first fact says capability corresponds to perplexity, so we can evaluate models by their cost to obtain it; the second says pretraining laws can be observed at the small scale; and the third says the final scale-up is the easy part.

We expect these assumptions to hold, but should detect when they do not. The shape of small-scale experiments lets us do just that: we can exploit the anticipated structure, check when it fails, and turn those failures into useful diagnostics---for example, if either the noisy quadratic limit or scaling law fails to emerge then there is likely an issue with the hyperparameters, model, or implementation.

More than checking assumptions, however, these diagnostics build a qualitative understanding of how the loss surface scales. Such qualitative understanding is critical because quantitative approaches can compound small errors, even when based on sound structure. More than pure extrapolation, a holistic approach is necessary.

\subsection{The Pitfalls of Pure Extrapolation}
\label{sec:designing-small-scale-experiments:the-pitfalls-of-pure-extrapolation}

\begin{wrapfigure}{r}{0.48\textwidth}
    \centering
    \vspace{-1.3cm}
    \includegraphics[width=0.43\textwidth]{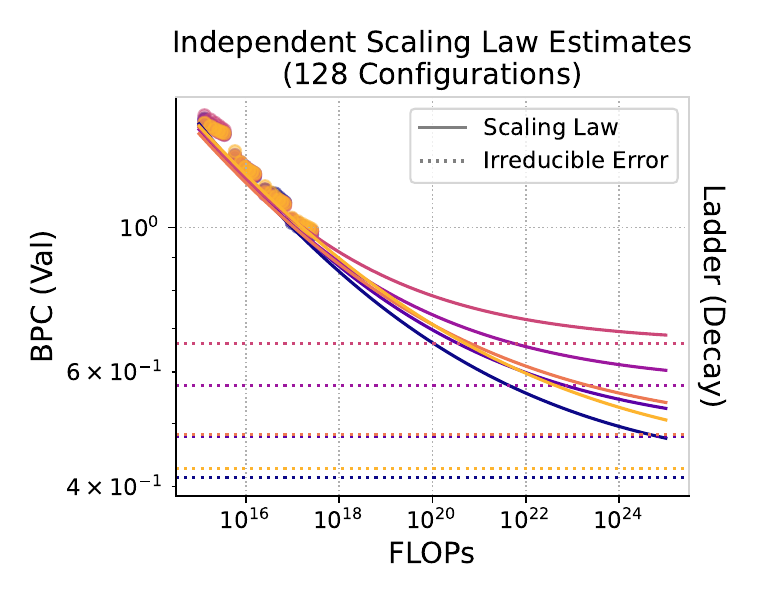}
    \caption{
        \emph{Despite good agreement near the data, extrapolation magnifies small differences due to sampling error.} We compare independent scaling law estimates, each based on 128 runs at 4M, 8M, 17M, and 34M parameters. Extrapolation eventually reflects the irreducible error, a quantity whose estimates vary greatly across different samples.
    }
    \label{fig:the-pitfalls-of-extrapolation}
    \vspace{-0.9 cm}
\end{wrapfigure}

Though scaling laws exist at small scales, their ability to extrapolate is another matter. We saw small-scale laws successfully extrapolate across an order of magnitude (\S\ref{sec:estimating-small-scale-scaling-laws}); however, what happens when we go further? \Cref{fig:the-pitfalls-of-extrapolation} 
% \sanae{update this if we add other panels to figure 1} 
illustrates how extrapolation magnifies the sampling error. Comparing independent estimates from different samples, we see the statistical noise: the irreducible error varies wildly across estimates.

If you compare models by extrapolating their scaling laws, at some point you are comparing their estimates of irreducible error. Such estimates are not reliable until the law begins to saturate. Scaling laws exist at small scales, and comparing them is useful; however, you cannot simply pick the model with the lowest final loss. The scaling law is there, but the statistical power is not. Instead, the law is much more reliable near the data.

\subsection{Case Study: Pre-Norm vs. Post-Norm Architectures}
\label{sec:designing-small-scale-experiments:case-study-pre-norm-vs-post-norm-architectures}

Where should normalization layers go in a transformer? The original placed them \emph{after} the residual connection (post-norm) \citep{Vaswani2017Attention}. This made training delicate, needing careful tuning. To stabilize training, \citet{baevski2018adaptiveinputrepresentationsneural} moved them \emph{before} the sublayer (pre-norm). A year later, an influential empirical study favored pre-norm in most settings yet, against today's practice, found post-norm better for high-resource machine translation \citep{nguyen-salazar-2019-transformers}. The matter only really settled when \citet{Xiong2020OnLayer} showed theoretically that pre-norm improves gradients at initialization. Pre-norm's advantage is most pronounced in deeper models, whose experiments are more expensive. Resolving it took nearly three years and a dedicated theoretical analysis. We will recover the answer from straightforward small-scale experiments.

A few well-separated scales should be far more informative than many nearby ones; we use three: two to fit the law, and one to validate it. For post-norm, we sample 511 configurations at 4M, 512 at 34M, and 128 at 134M; for pre-norm, we sample 128 at 4M, 34M, and 134M. We fit on the smaller two then evaluate on the largest one. This search is affordable because when parameters and data scale at the same rate compute is quadratic in model size. A single 1B-parameter run costs as much as 64 runs at 134M, 1{,}024 runs at 34M, or 65{,}536 runs at 4M. Thus, our experiment costs as much as a handful of billion parameter runs.

We now deepen our understanding of each architecture through a sequence of qualitative analyses. Each analysis is a diagnostic that asks a question and explores the structure we expect to find.

\paragraph{Diagnostic 1: have we tuned thoroughly?} If random search approaches the optimum, then the score distribution converges to a noisy quadratic as discussed in \Cref{sec:background:the-noisy-quadratic-limit} and established by \citet{lourie2025hyperparameter}. When the limit fails, it is likely the search missed the optimum.
% \footnote{
    % The noisy quadratic also enables deeper analyses such as confidence intervals for the best performance \citep{lourie2025hyperparameter}.
% } 
We visualize the limit for each architecture in \Cref{fig:noisy-quadratic-limit-for-prenorm-vs-postnorm}. For pre-norm it emerges readily; for post-norm it emerges too, but only deeper into the tail. This smaller asymptotic regime is an early indication of increased hyperparameter sensitivity. We seem to have located the optimum, but the tuning difficulty is itself our first finding: post-norm is sensitive to its hyperparameters, as the literature suggests.

\paragraph{Diagnostic 2: will scaling up be easy?} If larger models fail to become less sensitive to their hyperparameters, then transferring from small to large scales might fail as well. Therefore, we examine how sensitivity scales for the architectures at hand. \Cref{fig:sensitivity-across-scales-for-prenorm-vs-postnorm} compares the score distribution across scales for each. Both distributions move in the right direction, shifting mass towards the minimum as good configurations fill more of the space. For pre-norm the change is pronounced, growing a distinct peak at the optimum; for post-norm it is only slight, as a second peak lingers at a suboptimal value. Post-norm again exhibits tuning difficulty.

\paragraph{Diagnostic 3: does perplexity still track capabilities?} To evaluate by pretraining loss, it must correspond to capabilities (\S\ref{sec:background:perplexity-capability-correspondence}). This perplexity--capability correspondence should hold when pretraining data is fixed; still, the claim is easy to test as evaluating on a few tasks is much cheaper than pretraining the model. These tasks must show signal at small scales, like AI2 ARC (Easy) \citep{clark2018thinksolvedquestionanswering}. \Cref{fig:perplexity-capability-correspondence-for-prenorm-vs-postnorm} plots downstream accuracy vs. pretraining loss for several such tasks. Regardless of the architecture, all points fall on a common trend. Equal loss implies similar capabilities and a better pretraining law yields a more cost-effective model.

\paragraph{Diagnostic 4: do the scaling laws emerge?} To estimate the scaling laws, we adopt improvements from \S\ref{sec:estimating-small-scale-scaling-laws:the-impact-of-methodology}: tuning the hyperparameters per parameter--data budget and decaying the learning rate. Since extrapolation primarily reflects the noisy irreducible error (\S\ref{sec:designing-small-scale-experiments:the-pitfalls-of-pure-extrapolation}), we try fitting the laws jointly with tied irreducible error, as well as leaving them free to differ. \Cref{fig:validating-the-scaling-law} validates these scaling laws on the held-out scale. The free vs. joint laws extrapolate comparably. Pre-norm exhibits an excellent law, while post-norm's law is reasonable but could use improvement from further hyperparameter tuning. Again, hyperparameters present a challenge.

With a better understanding of each model, we now pool all sizes together and compare how the architectures scale. \Cref{fig:prenorm-vs-postnorm-compute-optimal-scaling} presents the result: pre-norm scales better over the range of sizes we explore. Remembering the statistical difficulty of extrapolating the scaling law (\S\ref{sec:designing-small-scale-experiments:the-pitfalls-of-pure-extrapolation}), we consider two kinds of comparisons. If we let each scaling have its own irreducible error, then the post-norm architecture appears to eventually outperform (beyond the scales we tested); however, if we share the irreducible error between the laws then the pre-norm architecture remains more efficient.

From a holistic viewpoint, post-norm was harder to tune at every turn. Even a better architecture in theory is worse in practice if we cannot find good hyperparameters. Beyond that, pre-norm scales better where the scaling laws are most reliable: near the data. Finally, when we assume both architectures reach the same final loss, pre-norm maintains its scaling advantage. These observations support pre-norm as the better choice.

%% file: contents/conclusion.tex
\section{Conclusion}
\label{sec:conclusion}

Scaling laws promised experiments would stay cheap; however, this promise has gone unfulfilled. Today, many believe regularity only emerges at large scales, and conclusions drawn at small scales will not transfer.
This misconception stems from a simple, unappreciated fact: small models are much more sensitive to their hyperparameters. While scaling laws extend to very small sizes, they only emerge on the fully tuned frontier. Reaching it requires an extensive search that researchers rarely conduct. Scaling laws predict pretraining loss, and better loss does not always mean better capability; however, loss also fully captures capability---as long as the data is held fixed. Therefore, a researcher who seeks not to predict capabilities but only to improve a model can rely on loss since a model with a better law buys every loss, and thus any capability, at lower cost.

These insights suggest a new methodology of small-scale experiments, one that exploits the strong assumptions we expect to hold true while checking when they do not. Working at small scales makes this rigor affordable: we can train hundreds of models then use statistical tools, like the noisy quadratic limit, to test our assumptions. Small-scale experiments save compute, but more than that they make our conclusions more robust.

% Limitation
\textbf{Nevertheless, small-scale experiments have several limitations.} While scaling laws exist at small scales, their extrapolation hits statistical challenges. The small scale is informative, but requires judgment. Looking at the bigger picture, is the model hard to tune? Is it getting easier? How does scaling compare where we trust the law most? Instead of a single test, we should explore each of these questions using different diagnostics.
Perhaps the most important limitation of all, though, is the restriction to model-centric research. Changing the data breaks the perplexity--capability correspondence which pretraining loss requires to proxy for downstream tasks. With neither pretraining loss as a proxy nor reliable downstream scaling laws, small-scale experiments have limited ability to predict large-scale outcomes. Hopefully future work will unlock small-scale experiments for data-centric research; still, with care we can use small-scale experiments for model-centric research today.

% Appendix
\begin{wrapfigure}{r}{0.44\textwidth}
    \centering
    \vspace{-0.9\baselineskip}
    \definecolor{plumdark}{HTML}{4A2472}
    \definecolor{plummed}{HTML}{7B5EA7}
    \definecolor{plumlight}{HTML}{C4A0C8}
    \begin{tikzpicture}
    \begin{axis}[
        name=cax,
        clip=false, scale only axis, width=0.30\textwidth, height=0.20\textwidth,
        xmin=0.4, xmax=4.6, ymin=0, ymax=80,
        axis y line=left, axis x line=bottom,
        xtick={1,2,3,4}, xticklabels={$u_1$,$u_2$,$u_3$,$u_4$},
        ytick={0,40,80}, tick label style={font=\tiny},
        ylabel={\tikz{\draw[draw=plummed, fill=plumlight, fill opacity=0.6] (0,0) rectangle (5pt,5pt);}\,$C_j^\star/C$ (\%)},
        ylabel style={font=\tiny, text=plummed},
        y tick label style={text=plummed},
        major tick length=0pt,
        ymajorgrids=true, major grid style={draw=gray!20, very thin},
    ]
      \addplot[ybar, bar width=9pt, draw=plummed, fill=plumlight, fill opacity=0.6]
          coordinates {(1,8.6) (2,1.1) (3,21.6) (4,68.7)};
      \draw[gray, dashed, thin] (axis cs:2.5,0) -- (axis cs:2.5,80);
    \end{axis}
    \begin{axis}[
        at={(cax.south west)}, anchor=south west,
        clip=false, scale only axis, width=0.30\textwidth, height=0.20\textwidth,
        xmin=0.4, xmax=4.6, ymin=0, ymax=36,
        axis y line=right, axis x line=none, xtick=\empty,
        ytick={0,15,30}, tick label style={font=\tiny},
        ylabel={\tikz{\node[circle, fill=plumdark, inner sep=0pt, minimum size=4pt]{};}\,$M_j^\star$},
        ylabel style={font=\tiny, text=plumdark},
        y tick label style={text=plumdark},
        major tick length=0pt,
    ]
      \addplot[plumdark, very thick, mark=*, mark size=1.8pt]
          coordinates {(1,32) (2,1) (3,5) (4,4)};
    \end{axis}
    \end{tikzpicture}
    \caption{
        \textbf{Getting the best scaling law fit requires different number of hyperparameter runs per scale.} The plot illustrates a typical solution, to optimizing the error at a scale $u_\star$  for a given compute budget. Because runs are cheap at $u_1$, we can run many runs while spending very little of the overall compute. The opposite is the case for the larger scale $u_4$. Another factor is the proximity to the center of the scales (dotted line). The closer a point is to the center the less influence it is going to have on the scaling law estimate, hence we are willing to make a larger mistake (see $u_2$). For the full derivation of this picture see \Cref{sec:compute-optimal-tuning}.
    }
    \label{fig:conclusion-compute-allocation}
    \vspace{-0.93\baselineskip}
\end{wrapfigure}

\textbf{Knowing what we know now, what would we change about how we design the experiments that estimate scaling laws?} Appendix~\ref{app:additional-theoretical-results} works through this question in full; here we sketch the reasoning.

First, the goal is not necessarily the most accurate scaling law, but the smallest error in the loss estimate $\hat{y}(u_\star)$ at the scale $u_\star$ we extrapolate to. And as we have shown, the two are not necessarily the same (see \Cref{fig:comparing-best-practices} for empirical and \Cref{sec:scaling-law-error-propagation} for theoretical evidence). 
A scaling law fit is similar to a linear regression in log-log space; in the linear case, we can propagate any error $\epsilon_j$ made at a small scale $u_j$ to its effect on the extrapolated prediction $\hat{y}(u_\star)$. We find that not all errors are equal, some will have a more severe effect on the prediction than others (see \Cref{fig:leverage}).
We then tie the error at each scale to the tuning effort spent there. With $M_j$ random-search runs at scale $u_j$, the expected error shrinks as $\epsilon_j \sim M_j^{-2/\gamma_j}$, where $\gamma_j$ is the effective number of hyperparameters at that scale (\Cref{app:background:how-close-does-random-search-get}). 

Combining both insights turns experiment design into a constrained optimization problem: given a fixed compute budget, we determine the number of runs at each scale to minimize the expected error at the target scale $u_\star$ (\Cref{sec:compute-optimal-tuning}). 
We can solve this problem in closed form. An illustration of what a solution will typically look like is in \Cref{fig:conclusion-compute-allocation}. 
 We develop this argument in full in Appendix~\ref{app:additional-theoretical-results} and distill it into a manual for designing small-scale experiments requiring the practitioner to only know their compute budget and the compute it takes to run a model at any given scale. However, executing this protocol we leave for future work.

%% file: contents/statements_and_contributions.tex
\subsection*{AI use statement}

We used AI assistants to help draft the text of this paper. In addition, we initially developed the theoretical models presented in Appendix~\ref{app:interpretable-models-of-the-hyperparameter-loss-surface} without AI assistance, then afterwards discussed these models, their justification, and possible extensions with AI assistants. Claude suggested the insight that any decent hyperparameters should drive a neural networks' loss to the irreducible error in the infinite parameters/data limit. In addition, some of the formulas used in implementing these models were derived and/or checked with AI assistance, in addition to being reviewed by the authors. AI coding assistants were used to help implement and run some of the experiments. Most plots were created with extensive help from AI assistants.

We take responsibility for the final content of this work,
including text, claims or artifacts produced with the aid of generative AI.

% \subsection*{Ethics statement}

%\subsection*{Reproducibility statement}

%\subsubsection*{Author Contributions}

\subsection*{Acknowledgments}

This work was supported in part through the NYU IT High Performance Computing resources, services, and staff expertise. We thank NYU HPC for their generous support and help in computational simulations and experiments. This work was supported by the Institute of Information \& Communications Technology Planning \& Evaluation (IITP) with a grant funded by the Ministry of Science and ICT (MSIT) of the Republic of Korea in connection with the Global AI Frontier Lab International Collaborative Research. We also thank He He for her invaluable feedback early on in this project, and Andrey Gromov for his insightful comments.

%% file: contents/appendix/appendix_related_work.tex
\section{Additional Related Work}
\label{sec:related-work}

\subsection{Compute Scaling}\label{app:scaling-laws}

\Cref{eq:scaling-law_joint}  omits the hyperparameters, describing loss at the \emph{fully tuned frontier}: the best attainable loss given the parameters and data 
% (Figure~\ref{fig:scaling-laws-at-the-fully-tuned-frontier})
; however, parameters and data do not vary independently. The model's gradients must be computed for every example, so their product determines the total compute: $c \propto p d$.
Given a budget, you face a trade-off between larger models and longer training. Navigating this trade-off is exactly why people turn to scaling laws. Minimizing \Cref{eq:scaling-law_joint} with fixed compute yields the condition:
\begin{equation}\label{eq:compute-optimality-condition}
    \iota \frac{\zeta}{p^\iota} = \kappa \frac{\eta}{d^\kappa}
\end{equation}
Hence the optimal parameters and data become power laws in compute:
\begin{equation}\label{eq:compute-optimal-parameters-and-data}
    p \propto c^{\frac{\kappa}{\iota + \kappa}} \qquad d \propto c^{\frac{\iota}{\iota + \kappa}}
\end{equation}
If you plug these values into the joint scaling law then you obtain excess loss as a power law in the compute:
\begin{equation}\label{eq:scaling-law_compute}
    \mathcal{L}(p, d) - \epsilon \propto c^{-\frac{\iota\kappa}{\iota + \kappa}}
\end{equation}
Thus scaling laws reveal the best loss given the compute and how to get it by balancing parameters and data.

\citet{Hoffmann2022AnEmpirical} found balanced parameters and data grow at the same rate, a conclusion since echoed by many others \citep{anil2023palm2technicalreport, deepseekai2024deepseekllmscalingopensource}. This idea justifies the token-to-parameter ratio as a meaningful measure of how well-trained a model is, with the compute-optimal ratio serving as baseline. Under \Cref{eq:compute-optimality-condition}, this same-rate growth only occurs when the scaling exponents are equal: $\iota = \kappa$. Some theoretical work predicts this equality \citep{Bahri2024Explaining}, and the estimates are often close in practice. Researchers have also explored enforcing it to improve estimation \citep{Muennighoff2025Scaling, li2025misfitting}.

\subsection{The Noisy Quadratic Limit}
\label{app:background:the-noisy-quadratic-limit}

How do we know if the hyperparameters are fully tuned, or how hard that is to accomplish? The \emph{noisy quadratic limit} answers these questions. \citet{lourie2025hyperparameter} found that near the optima, loss surfaces are:
\begin{enumerate}[noitemsep]
    \item \textbf{Quadratic}: well approximated by the second-order Taylor expansion;
    \item \textbf{Low-Rank}: governed by a few directions that explain most of the variation; and
    \item \textbf{Normal}: distributed about the mean as additive normal noise.
\end{enumerate}
Empirically, this approximation holds for a large region around the optimum, the \emph{asymptotic regime}. This region determines the practical questions we care about, like tuning difficulty or the best attainable performance.

Formally, if $\bX$ are the hyperparameters, $Y$ the validation score, and $(\bx_*, y_*)$ the optimum, then the conditional distribution $Y\mid\bX$ is approximately a quadratic plus additive normal noise, $E$.\footnote{
    Integral hyperparameters (e.g., batch size) can be treated as continuous; discrete ones are handled with separate quadratics.
} Letting $H_{\bx_*}$ be the Hessian:
\begin{equation}\label{eq:noisy-quadratic-structure}
    \mathcal{L}(\bX) \approx y_* + \left(\bX - \bx_*\right)^T H_{\bx_*} \left(\bX - \bx_*\right) + E, \qquad E \sim \normal(0, \sigma)
\end{equation}

This conditional view induces a marginal structure that emerges during random search. More precisely, the search induces a distribution over validation scores, and that distribution's tail converges to a \textit{noisy quadratic distribution}, $\quadratic_{\min}(\alpha, \beta, \gamma, \sigma)$. Writing $F(y; \alpha, \beta, \gamma, \sigma)$ for the CDF of the noisy quadratic distribution, then:
\begin{equation}\label{eq:noisy-quadratic-limit}
    \Prob(Y \leq y) \approx F(y; \alpha, \beta, \gamma, \sigma) \quad\text{ as }\quad y \to y_*.
\end{equation}
The distribution's parameters capture properties of the loss surface: $\alpha$ is the best attainable performance ($y_*$), $\beta$ measures how tightly the scores concentrate around it, $\gamma$ gives the \textit{effective} number of hyperparameters (intrinsic dimension at the optimum), and $\sigma$ represents the noise due to random seeds. In practice, we define the asymptotic regime by a threshold on the expected loss, $\E[Y\mid\bX] \leq \theta$, marking where this structure holds.

The \textbf{noisy quadratic distribution} arises as the sum $Q + E$, where $Q \sim \quadratic_{\min}(\alpha, \beta, \gamma)$ is the (noiseless) quadratic distribution and $E \sim \normal(0, \sigma)$ is independent Gaussian noise. The noiseless quadratic is supported on $[\alpha, \beta]$ with CDF and density
\begin{equation}\label{eq:quadratic-cdf}
    F_Q(y; \alpha, \beta, \gamma) = \left(\frac{y - \alpha}{\beta - \alpha}\right)^{\gamma/2},
    \qquad
    f_Q(y; \alpha, \beta, \gamma) = \frac{\gamma}{2(\beta - \alpha)}\left(\frac{y - \alpha}{\beta - \alpha}\right)^{\frac{\gamma - 2}{2}},
\end{equation}
for $\alpha \leq y \leq \beta$ (and $0$, resp. $1$, outside). Its lower tail decays polynomially with exponent $\gamma/2$, encoding the ellipsoidal volume of near-optimal hyperparameters: the rank of the Hessian sets $\gamma$, so higher intrinsic dimension makes near-optimal configurations rarer. Convolving with the Gaussian gives the noisy quadratic, supported on all of $\reals$, with CDF and density
\begin{align}\label{eq:noisy-quadratic-cdf}
    F(y; \alpha, \beta, \gamma, \sigma)
        &= \Phi\!\left(\frac{y - \beta}{\sigma}\right)
            + \int_{0}^{1} v^{\gamma/2}\, p_V(v)\, \d v,
    \\ \label{eq:noisy-quadratic-pdf}
    f(y; \alpha, \beta, \gamma, \sigma)
        &= \frac{\gamma}{2(\beta - \alpha)} \int_{0}^{1} v^{\frac{\gamma - 2}{2}}\, p_V(v)\, \d v,
\end{align}
where $\Phi$ is the standard normal CDF and $p_V$ is the density of $V \sim \normal\!\left(\frac{y - \alpha}{\beta - \alpha}, \frac{\sigma}{\beta - \alpha}\right)$. 
The $\Phi$ term accounts for noise carrying scores past $\beta$, while the truncated integral convolves the quadratic tail with the noise over the support $[\alpha, \beta]$ (i.e., $v \in [0, 1]$). 
As $\sigma \to 0$, the integral concentrates and \Cref{eq:noisy-quadratic-cdf} recovers the noiseless quadratic \Cref{eq:quadratic-cdf}.

\subsection{Downstream Scaling laws}
\label{sec:appendix-Downstream-Scaling-laws}

\begin{wrapfigure}{r}{0.7\textwidth}
    \centering
    \includegraphics[width=0.95\linewidth]{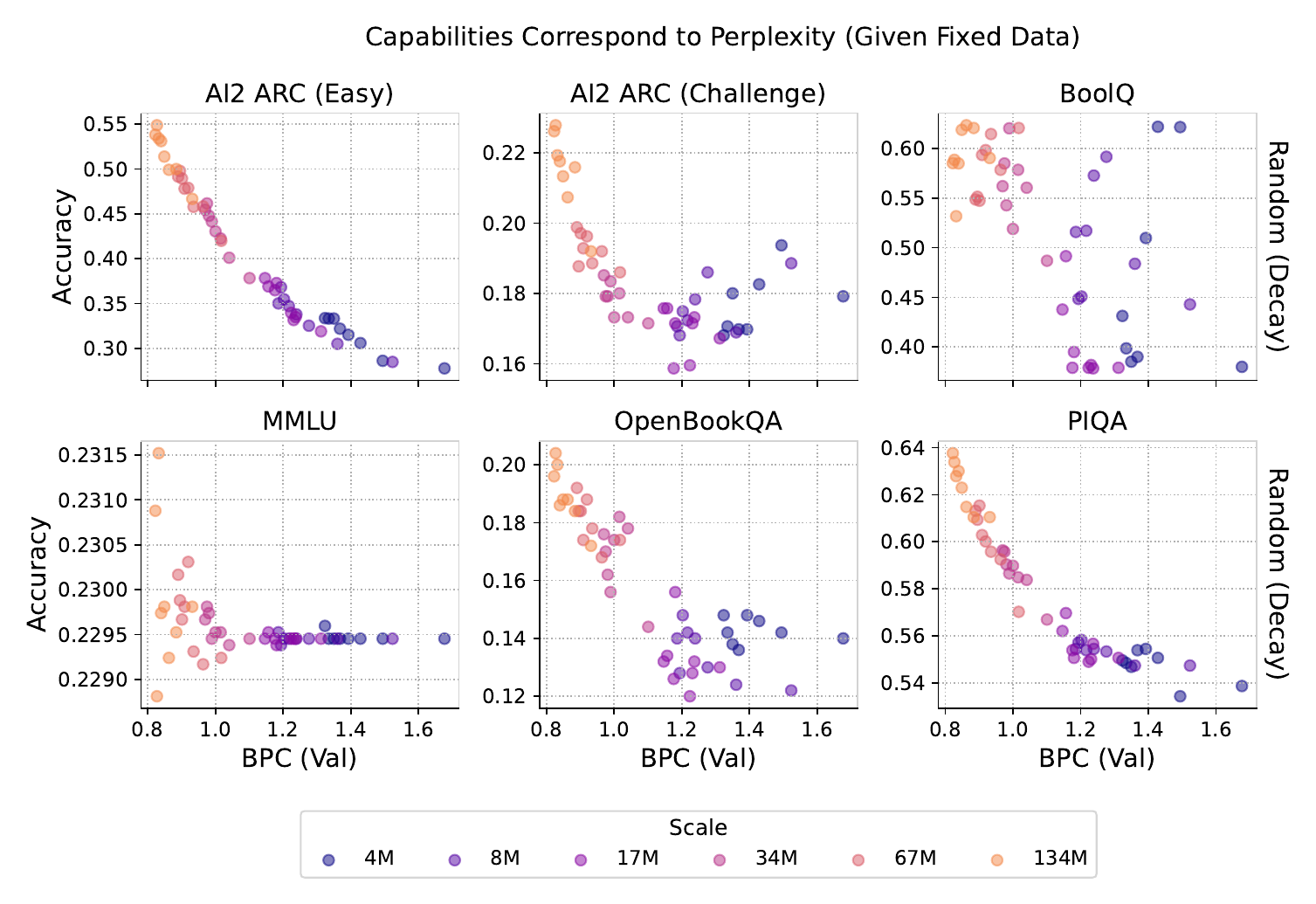}
    \caption{
        \emph{Given a fixed corpus, perplexity corresponds to downstream capability across a wide variety of tasks.} The relationship between pretraining loss and capability is not always predictable or even monotonic; however, models achieving the same loss have a similar conditional distribution over their capabilities as we increase parameters and data, or even randomize the architecture, as long as the \emph{composition} of pretraining data remains fixed.
    }
    \label{fig:perplexity-corresponds-to-capabilities}
\end{wrapfigure}

The data defines a curve of capabilities; the loss decides a point on that curve; the scaling law determines the cost of that point. \Cref{fig:perplexity-corresponds-to-capabilities} illustrates this principle: models trained at different scales or for different lengths fall along a single (possibly noisy) curve. The curve varies wildly across tasks, but exists for all of them. Regardless of whether better capabilities lie down the road, the scaling law sets the toll. In other words a model with a better scaling law achieves every perplexity, and thus any capability, at lower cost.

%% file: contents/appendix/experimental_design.tex
\section{Experimental Design}
\label{app:experimental-design}

Our experiments explore the hyperparameter loss surface across scales via random search.

\paragraph{Models.} We measure model size by the \textit{effective parameter count}, parameters adjusted for FLOPs, so that compute is $c = 6pd$ for $p$ effective parameters and $d$ tokens. This count excludes the embedding layer (which uses no FLOPs) but includes the unembedding layer and the cost of attention \citep[\S B]{Porian2024Resolving}. We target parameter counts at increasing powers of two: $2^{22}\approx4\text{M}$ to $2^{28}\approx268\text{M}$, and construct models at each scale with two experimental designs: a hand-crafted model ladder and randomly sampled architectures.

\paragraph{Training and evaluation.} We train with a warmup-stable-decay (WSD) schedule \citep{hu2024minicpm}. Following \citet{Hagele2024ScalingLaws}, we reuse the stable phase from the main run and branch off decay phases from checkpoints taken $\sfrac{1}{8}, \sfrac{2}{8}, \ldots, \sfrac{8}{8}$ through training. We evaluate checkpoints before and after decaying the learning rate, thus a single sweep yields stable and decayed evaluations across eight token budgets at a fraction of the cost.

\paragraph{Analysis.} We partition the scales into training (4M--34M), validation (67M--134M), and test (268M), so reported numbers measure extrapolation from small models to larger held-out ones. By default, we fit the joint scaling law of Equation~\ref{eq:scaling-law_joint} to the best loss attained at each parameter--data budget pair. Like prior work \citep{hilton2023scalinglawssingleagentreinforcement, li2025misfitting}, we found shorter training runs fit the scaling law poorly, therefore we discard the two smallest token budgets which noticeably improves performance on our validation set.

\subsection{Modeling}
\label{app:experimental-design:modeling}

The modeling recipe we study is a Llama variant, as implemented in Meta Lingua: \url{https://github.com/facebookresearch/lingua} (commit: 437d680) \citep{meta-lingua}. Hyperparameters not specified here were either left to Lingua's defaults or swept over as described in \Cref{app:experimental-design:search-distributions}.

\paragraph{Data.} We pretrain on the 100B token subset of FineWeb-Edu \citep{Penedo2024TheFineWeb}. We use the subset to save storage and preprocessing since the scales we consider require less than 100B tokens. We split this subset into training and validation sets using Lingua's built-in preprocessing pipeline with 48GB of memory and Lingua's suggested seed of 42. We create two such datasets: one shuffling all the data in 1 chunk for single GPU training (4M--134M parameters), and another splitting the data into 8 chunks for 8 GPU training (268M parameters).

\paragraph{Architecture.} For the tokenizer, we use \texttt{p50k\_base} from \texttt{tiktoken} (\url{https://github.com/openai/tiktoken}), the BPE tokenizer from GPT-3.5 (\texttt{text-davinci-003}), because it has an appropriately sized vocabulary for our scale. The context length is 1{,}024 with special tokens to distinguish the beginning and ending of each sequence. For the neural network, we use a decoder-only Llama model \citep{grattafiori2024llama3herdmodels} available in Lingua (\texttt{apps/main/train.py}) \citep{meta-lingua}. The model uses RMSNorm for normalization \citep{Zhang2019Root}, SwiGLU for the activation function \citep{shazeer2020gluvariantsimprovetransformer}, and RoPE for position embeddings \citep{Su2024RoFormer}. All runs use a pre-normalization architecture except for those explicitly specified as post-normalization (e.g., in \S\ref{sec:designing-small-scale-experiments}), which we implemented as described by \citet{Xiong2020OnLayer}. Aside from Lingua's defaults, we padded the vocabulary size (\texttt{model.vocab\_size}) to 50{,}536. In addition, we set \texttt{model.multiple\_of} to 8 since the default value of 256 is larger than our smaller models' hidden dimensions; 8 compromises between allowing a range of hidden dimensions at small scales while partially exploiting increased GPU efficiency.

\paragraph{Optimization.} For optimization, we use AdamW \citep{loshchilov2018decoupled} with a warmup-stable-decay (WSD) learning rate schedule \citep{hu2024minicpm}. Unlike Lingua, we use a linear cooldown for the WSD schedule decaying to a \texttt{optim.lr\_min\_ratio} of 1e{-}6 times the peak learning rate. Each scale we train is defined by a target number of effective parameters (described in \Cref{app:experimental-design:search-distributions}), ranging from $2^{22} \approx 4M$ to $2^{28} \approx 268M$ in powers of two. Training continues until the number of tokens equals 32 times the target number of effective parameters. The total tokens divided by the tokens per gradient step determines the number of steps. The effective parameter count is essentially the same as FLOPs per step as used in \citet{Hoffmann2022AnEmpirical} (Approaches 1 \& 2), thus we train about 50\% further than the Chinchilla-optimal ratio of $\sim$20 tokens-per-parameter, allowing us to explore both the under- and over-trained regimes. Following \citet{Hagele2024ScalingLaws}, we first train each configuration on a main run without decaying the learning rate, saving checkpoints $\sfrac{1}{8}, \sfrac{2}{8}, \ldots, \sfrac{8}{8}$ through training; afterwards we decay each of those checkpoints by extending training with 25\% more steps, thus the final results have their learning rates decayed over the last 20\% of training. The initial checkpoints provide our constant learning rate results (\emph{stable}), whereas the decayed checkpoints provide our results for the full schedule (\emph{decay}); all results include learning rate warmup.

\paragraph{Evaluation.} For pretraining, we evaluate the model via bits-per-character (BPC), which normalizes the pretraining loss for differences in tokenization. To compute BPC, we change the logarithm's base on the negative log-likelihood per character already provided by Lingua. We calculate BPC on the validation set, limiting \texttt{eval.validation.max\_steps} to 1{,}000. For capabilities, we evaluate the model using Lingua's integration with the EleutherAI \texttt{lm-evaluation-harness} (v0.4.11) \citep{eval-harness}. We evaluate on six multiple-choice tasks: AI2 ARC (Easy) and (Challenge) \citep{clark2018thinksolvedquestionanswering}, BoolQ \citep{clark-etal-2019-boolq}, MMLU \citep{hendrycks2021measuring}, OpenBookQA \citep{mihaylov-etal-2018-suit}, and PIQA \citep{Bisk2020PIQA}. Our evaluations limit \texttt{eval.generator.max\_tokens} to 16{,}384 but otherwise use Lingua's defaults.

\paragraph{Random Seeds.} We fix the following random seeds made available by Lingua: \texttt{seed} to 0, \texttt{data\-.seed} to 1, \texttt{model\-.seed} to 2, \texttt{eval\-.harness\-.random\_seed} to 3, \texttt{eval\-.harness\-.numpy\_random\_seed} to 4, \texttt{eval\-.harness\-.torch\_random\_seed} to 5, \texttt{eval\-.harness\-.fewshot\_random\_seed} to 6. Accordingly, data order and other details were kept constant while runs differed primarily in their hyperparameters.

\subsection{Search Distributions}
\label{app:experimental-design:search-distributions}

Our results analyze three different experiments, each defined by a search distribution over the hyperparameters.

All the experiments randomly sample configurations across a range of scales, with each scale defined by the \emph{effective} number of parameters. The effective number of parameters is proportional to the FLOPs per token so that $c = 6pd$ when $c$ is compute, $p$ the effective parameters, and $d$ the number of tokens. Following \citet[\S B]{Porian2024Resolving}, we compute it as:
\begin{equation}\label{eq:effective-number-of-parameters}
    N_{eff} \coloneqq \left(3 d_{ff} + 4d\right)dl + dv + ndl
\end{equation}
where $d_{ff}$ is the feedforward dimension, $d$ the hidden dimension, $l$ the number of layers, $v$ the vocabulary size, and $n$ the context length of the transformer. In particular, parameters in the embedding layer are not counted because they do not contribute FLOPs, while parameters in the unembedding layer are counted because they do. See \citet[\S B]{Porian2024Resolving} for an overview of different ways to count parameters.

\begin{figure}
    \centering
    \begin{align*}
        \hparam{batch\_size}   &\sim \distribution{DiscreteUniform}\left(\{64, 128, 256, \ldots, 4{,}096\}\right) \\
        \hparam{lr}            &\sim \distribution{LogUniform}(1e{-}5, 1e{-1}) \\
        \hparam{beta1}         &\sim \distribution{LogitUniform}(0.7, 0.999) \\
        \hparam{beta2}         &\sim \distribution{LogitUniform}(0.9, 0.9999) \\
        \hparam{warmup}        &= \operatorname{round}\left(\hparam{steps} \times P\right),\quad P \sim \distribution{LogUniform}\left(1e{-}3, \sfrac{1}{8}\right) \\
        \hparam{weight\_decay} &\sim \distribution{LogUniform}(1e{-}4, 1e0) \\
        \hparam{rope\_theta}   &\sim \distribution{LogUniform}\left(2^{10}, 2^{20}\right) \\
    \end{align*}
    \caption{
        \emph{The search distribution for all non-architectural hyperparameters.} The log uniform and logit uniform distributions are obtained by taking the log or logit of the bounds, sampling uniformly on this scale, then inverting the transformation back to the original scale.
    }
    \label{fig:search-distribution_non-architectural}
\end{figure}

To isolate the effect of tuning the architecture, configurations of the same index share all non-architectural hyperparameters across the different designs. Thus, all differences between the experiments come from the architecture alone. The search distribution over the shared hyperparameters is given in \Cref{fig:search-distribution_non-architectural}. Typically, researchers search hyperparameters like the learning rate or weight decay on a log scale; using the proper scale for each hyperparameter is also important in efficiently reaching the noisy quadratic limit \citep{lourie2025hyperparameter}. In practice, we found the simple heuristic of sampling real values on a linear scale, positive values on a log scale, and bounded values on a logit scale worked quite well. The warm up is sampled as a proportion of the total training steps up to the first $\sfrac{1}{8}$ of training so that the first checkpoint (and evaluation) is always fully warmed up.

\begin{table}
    \centering
    \begin{tabular}{c|rrrr}
        \toprule
          $N_{eff}$ & $\hparam{dim}$ & $\hparam{n\_layers}$ & $\hparam{head\_dim}$ & $\hparam{n\_heads}$ \\
        \midrule
              3.47M &             64 &                    2 &                   64 &                   1 \\
              7.78M &            128 &                    4 &                   64 &                   2 \\
              17.1M &            256 &                    4 &                   64 &                   4 \\
              32.4M &            384 &                    6 &                   64 &                   6 \\
              62.6M &            512 &                   10 &                   64 &                   8 \\
               133M &            768 &                   12 &                   64 &                  12 \\
               270M &          1,024 &                   16 &                   64 &                  16 \\
        \bottomrule
    \end{tabular}
    \caption{
        \emph{The model ladder used for architectural hyperparameters in the \texttt{ladder} experiment.} The leftmost column gives the actual effective number of parameters which differs slightly from the targeted number due to architectural constraints.
    }
    \label{tab:model-ladder}
\end{table}

\paragraph{Model Ladder (\texttt{ladder}).} We sample configurations for target scales starting from $N_{eff} = 2^{22} \approx 4M$ up to $2^{28} \approx 268M$ in powers of two. Excluding 2 failed runs, we search 867 configurations at 4M effective parameters, 826 at 8M, 867 at 17M, 934 at 34M, 128 at 67M, 128 at 134M, and 64 at 268M. For the \texttt{ladder} experiment, we follow the common practice of setting architectural hyperparameters via a hand-crafted model ladder \citep{Hagele2024ScalingLaws, bhagia2025establishing}. The hidden dimension, number of layers, and number of attention heads at each scale were set according to Table~\ref{tab:model-ladder}. We increase the width and depth of successive models in order to steadily increase their parameter count, while keeping other architectural elements (such as the $\hparam{head\_dim}$) constant. Following \citet{Porian2024Resolving}, we keep the aspect ratio between 32 and 64. The transformer architecture is subject to a number of constraints, depending on the implementation (e.g., $\hparam{head\_dim}$ divides $\hparam{dim}$, $\hparam{dim}$ is used both for the hidden and the embedding dimension, matrix dimensions should be multiples of 8 when running on a GPU, and so on). These constraints must be considered when designing a ladder.

\begin{figure}
    \centering
    \begin{align*}
        \hparam{n\_layers} &= \operatorname{round}\left(U^2\right),\quad U \sim\distribution{Uniform}\left(\sqrt{2}, \sqrt{48}\right) \\
        \hparam{n\_heads}  &\sim \distribution{DiscreteProportional}\left(\left\{k \mid k \text{ divides } \sfrac{\hparam{dim}}{8},\; 1 \leq k < \min\left(24, \sfrac{\hparam{dim}}{8}\right)\right\}\right) \\
    \end{align*}
    \caption{
        \emph{The search distribution for architectural hyperparameters used in the \texttt{random}, \texttt{prenorm}, and \texttt{postnorm} experiments.} The discrete proportional distribution samples from a finite set of positive numbers with probability proportional to each number. By construction, $\hparam{dim}$ is always divisible by $\text{\texttt{model.multiple\_of}} = 8$.
    }
    \label{fig:search-distribution_architectural}
\end{figure}

\paragraph{Random Architecture (\texttt{random}).} We sample configurations for target scales starting from $N_{eff} = 2^{22} \approx 4M$ up to $2^{27} \approx 134M$ in powers of two. Excluding 2 failed runs, we search 128 configurations at 4M effective parameters, 126 at 8M, 128 at 17M, 128 at 34M, 128 at 67M, and 128 at 134M. In the \texttt{random} experiment, we also tune the architectural hyperparameters via random search. The number of layers and attention heads for each configuration are sampled from the distribution described in \Cref{fig:search-distribution_architectural}. Given the effective parameters, the number of layers determines the hidden dimension via \Cref{eq:effective-number-of-parameters}, which is a quadratic in $d$ given the other parameters and the fact that $d_{ff} = \sfrac{8}{3} d$ for our model. We round the solution to the nearest multiple of 8 for GPU efficiency. Once we have the hidden dimension, the number of heads determines the head dimension: $\hparam{dim} = \hparam{head\_dim} \times \hparam{n\_heads}$. We ensure $\hparam{n\_layers}$ is at least 2 because induction heads cannot form with fewer layers \citep{olsson2022incontextlearninginductionheads}; we also use a square root scale because we found it balanced making both the number of layers and the hidden dimension more uniform. $\hparam{n\_heads}$ must divide the hidden dimension, so we sample proper divisors of $\sfrac{\hparam{dim}}{8}$ to guarantee that $\hparam{head\_dim}$ is even (required by RoPE) and greater than 16 (required by the attention implementation). Since small divisors are much more common, we bias towards larger ones by sampling divisors with probability proportional to their size. As previously mentioned, transformers are subject to a number of constraints that must be accounted for when sampling architectural hyperparameters: $\hparam{head\_dim} \geq 16$ for common attention implementations, $\hparam{head\_dim}$ must be even for RoPE, $\hparam{head\_dim}$ divides $\hparam{dim}$, $\hparam{dim}$ provides both the hidden and the embedding dimension, and matrix dimensions should be multiples of 8 for GPU efficiency. These constraints complicate designing a good search distribution;\footnote{
    A typical model ladder avoids most of these complications, as they primarily appear as edge cases in small networks.
} the distribution in \Cref{fig:search-distribution_architectural} has been constructed to obey them.

\paragraph{Pre-Normalization (\texttt{prenorm}) vs. Post-Normalization (\texttt{postnorm}).} We sample configurations for target scales of $N_{eff} = 2^{22} \approx 4M$, $2^{25} \approx 34M$, and $2^{27} \approx 134M$. For \texttt{prenorm}, we reuse all the results at those scales from the \texttt{random} experiment. For \texttt{postnorm}, we reuse the initial configurations of the \texttt{random} experiment (including architectural parameters), but we re-run them after changing the transformer to a post-normalization architecture. In addition, we expand the number of configurations due to the post-normalization architecture's increased tuning difficulty. Excluding 1 failed run, this expansion results in 511 configurations at 4M effective parameters, 512 at 34M, and 128 at 134M.

\subsection{Analysis}
\label{app:experimental-design:analysis}

\paragraph{Noisy Quadratic Limit.} We generate confidence bands for the eCDF using the LD Highest Density bands from \citet{lourie-etal-2024-show}, and we fit the noisy quadratic distribution as described in \citet{lourie2025hyperparameter}, both using \texttt{opda} (v0.8.0): \url{https://github.com/nicholaslourie/opda}. To fit the noisy quadratic limit, we set the threshold $\theta$ at various percentiles of the score distribution. We used different percentiles for each parameter--data budget, with the size of the asymptotic regime generally increasing in both. In the \texttt{ladder} and \texttt{random} / \texttt{prenorm}) experiments, they were as follows: for 4M, 8M, and 17M effective parameters, 7\% at checkpoints 1--6 and 8\% at checkpoints 7--8; for 34M, 7\% for checkpoints 1--2, 8\% at 3, 10\% at 4--6, and 12\% at 7--8; for 67M, 12\% at 1, 15\% at 2, 20\% at 3, 25\% at 4, 30\% at 5, 32\% at 6, and 35\% at 7--8; for 134M and 268M, 12\% at 1, 15\% at 2, 20\% at 3, 25\% at 4, 30\% at 5, 35\% at 6, 40\% at 7, and 50\% at 8. In the \texttt{postnorm} experiment, they were as follows: for 4M effective parameters, 3\% at checkpoint 1, 4\% at 2, 5\% at 3, 6\% at 4, and 7\% at 5--8; for 34M and 134M, 7\% at 1--2, 8\% at 3, and 10\% at 4--8. We restricted $\alpha$ to be positive (as bits-per-character always is), and we restricted $\gamma$ to be 1 to at most the nominal number of hyperparameters (7 for \texttt{ladder} and 9 for \texttt{random} / \texttt{prenorm} and \texttt{postnorm}).

\paragraph{Scaling Laws.} To fit and evaluate the scaling laws, we partition the scales into training (4M--34M), validation (67M--134M), and testing (268M); thus, we evaluate the scaling laws based on their ability to extrapolate from small to large models. We used training to fit the laws, validation to make decisions during our research, and testing for the final evaluation. For the scaling law, we fit \Cref{eq:scaling-law_joint} to the best validation loss from each parameter--data budget pair, or (where explicitly stated) to the validation losses from the run achieving the minimum final loss at each scale. In both cases, we drop the first two checkpoints ($\sfrac{1}{8}$ and $\sfrac{2}{8}$ of training) as it substantially improved the fit \citep{hilton2023scalinglawssingleagentreinforcement}. Where specified, we also tie the scaling exponents together ($\iota = \kappa$). To perform the optimization, we minimized the mean squared error via SciPy's \texttt{scipy.optimize.differential\_evolution} \citep{SciPy2020Virtanen}, with bounds of $[0, 1]$ for $\epsilon$, $[0, 10]$ for $\ln \zeta$ and $\ln \eta$ (which we fit on log scales), and $[0, 1]$ for $\iota$ and $\kappa$. We used a \texttt{popsize} of 120 except when fitting laws jointly with a single irreducible error for which we used a \texttt{popsize} of 60 due to the higher dimensionality and longer runtime. To evaluate the scaling laws, we report mean squared error between the predicted and observed loss on the held-out scales, with held-out scales using the same preprocessing as those in training (e.g., whether the best run or the best configuration for each parameter--data budget is used).

\paragraph{Interpretable Models of the Hyperparameter Loss Surface.} In \Cref{app:interpretable-models-of-the-hyperparameter-loss-surface}, we compare various models of the hyperparameter loss surface. These models are fit to losses generated via the same methodology used for scaling laws (best loss for each parameter--data budget, dropping the first two checkpoints), except instead of selecting the loss for the single best configuration, we take all configurations within the asymptotic regime as defined by the fractions used for our noisy quadratic limits. Before fitting the models, we transform the hyperparameters to put them on an appropriate scale: a log-scale for $\hparam{n\_effective\_parameters}$, $\hparam{n\_tokens}$, $\hparam{batch\_size}$, $\hparam{lr}$, $\hparam{weight\_decay}$, $\hparam{dim}$, $\hparam{n\_layers}$, $\hparam{n\_heads}$, and $\hparam{rope\_theta}$; a log-plus-one-scale for $\hparam{warmup}$ (to avoid log of zero); and a logit-scale for $\hparam{beta1}$ and $\hparam{beta2}$. The interpretable models are fit by first taking a good initialization and then improving it with SciPy's \texttt{scipy.optimize.minimize} using the \texttt{"Nelder-Mead"}, \texttt{"Powell"}, and \texttt{"BFGS"} methods in sequence. For the initialization, we fit a quadratic polynomial via ordinary least squares, algebraically solve for the equivalent estimate in the format of \Cref{eq:static-empirical-model_projection}, and then project the Hessian to a lower rank ($\gamma = 3$) by dropping the eigenvalues of least absolute value. We then use the \texttt{static} model to initialize the \texttt{dynamic optima} and the \texttt{dynamic sensitivity} models, setting the new components to zero since the models are nested. We similarly initialize the \texttt{dynamic} model from the \texttt{dynamic optima} model. For the \texttt{multilayer perceptron} model, we implement the neural network using scikit-learn \citep{scikit-learn}. The pipeline applies a quadratic basis expansion to the hyperparameters, normalizes the features via \texttt{preprocessing.RobustScaler}, and finally predicts via a multilayer perceptron (MLP) with \texttt{hidden\_layer\_sizes=[128]}, \texttt{learning\_rate="adaptive"}, \texttt{batch\_size=256}, and \texttt{max\_iter=4096}. We grid search \texttt{alpha} over 1e{-}4, 1e{-}3, 1e{-}2, 1e{-}1, 1e0 and \texttt{learning\_rate\_init} over 1e{-}5, 3e{-}5, 1e{-}4, 3e{-}4, 1e{-}3 using the train-validation split described previously. For the MLP, we fit the scaling law beforehand and then fit the MLP to the percent excess loss, rather than fitting them jointly.

\subsection{Implementation}
\label{app:experimental-design:implementation}

Our implementation further involves the following details.\footnote{
    Preliminary experiments were run on the NYU Torch cluster. We later expanded our random searches using Meta's compute infrastructure. This section describes the compute setup used at NYU, which is sufficient to reproduce our findings.
}

\paragraph{Training Pipeline.} We use gradient accumulation so that we can tune the batch size despite different scales requiring different amounts of memory and GPU setups (1 vs. 8 GPUs). We set the per-GPU batch size to the largest power of two that safely avoids out-of-memory errors in order to maximize throughput while also evenly dividing the number of tokens (16 sequences for 4M--67M parameters, 8 sequences for 134M--268M parameters). Each scale is assigned a base number of gradient accumulation steps chosen so that after accounting for the per-GPU batch size and the number of GPUs the model will process 64 sequences per update. We then multiply that base by a power of two to achieve the distribution described in \Cref{fig:search-distribution_non-architectural} for the total batch size. As a result, the batch size sweep is hardware-agnostic and always evenly divides the total number of training tokens. In addition, we set \texttt{gc\_collect\_freq} to 1{,}024, \texttt{data.prefetch\_size} to 1{,}024, \texttt{distributed.compile} to \texttt{true}, and \texttt{distributed.fsdp\_type} to \texttt{"full\_shard"} in order to make training efficient. Finally, we set \texttt{logging.freq} to 1 and found it necessary to set \texttt{env.ENABLE\_INTRA\_NODE\_COMM} to \texttt{"0"} to avoid an unresolved error in the PyTorch version.

\paragraph{Software.} We train the models using Meta Lingua (commit: 437d680), as previously described \citep{meta-lingua}. We modify the code base to support the 100B token subset of FineWeb-Edu, use a linear decay in the WSD schedule, and use PyTorch's Scaled Dot Product Attention (SDPA) instead of Flex Attention as Flex Attention would not successfully generate kernels for some of the transformer shapes that we randomly sample. In addition, we switch the transformer architecture from pre-norm to post-norm for some experiments as previously discussed. We isolate the software environment in an Apptainer (Singularity) container \citep{singularity2021}. The container runs Ubuntu 24.04.2 LTS with CUDA 12.8 (V12.8.93) and cuDNN 9.8, and we run Lingua via a Python 3.11.14 conda environment with PyTorch 2.7.0+cu128 \citep{Paszke2019PyTorch}.

\paragraph{Hardware.} Training was done on 80GB NVIDIA A100 GPUs on a SLURM cluster. Model sizes from 4M--134M effective parameters were trained with 1 GPU, while models with 268M effective parameters were trained data-parallel with 8 GPUs.

%% file: contents/appendix/model_of_hp_surface.tex
\section{Interpretable Models of the Hyperparameter Loss Surface}
\label{app:interpretable-models-of-the-hyperparameter-loss-surface}

In \Cref{sec:understanding-the-hyperparameter-loss-surface-across-scales:implications-for-small-scale-experiments}, we discussed developing an interpretable model of the hyperparameter loss surface across scales by combining the noisy quadratic limit with scaling laws. We reported that such models capture the broad shape of the surface, but not precisely how it evolves. Here we provide those explorations in more depth. In particular, we will compare several approaches a researcher might be tempted to try and show that while they have some success in modeling the loss surface, they are far from the clear wins available via scaling laws or the noisy quadratic limit on their own.

Scaling laws model the optimal loss across scales; the noisy quadratic limit models how hyperparameters influence the loss at one scale. To combine them, we must set the quadratic's minimum according to the scaling law. One could accomplish this in several ways;\footnote{
    Beyond the approach discussed here, we also tried using the scaling law for the quadratic's constant term which is perhaps the most obvious approach even if it does not have an a priori justification.
} a simple approach with a strong intuition starts from the observation that, given any reasonable hyperparameters, the loss should approach the irreducible error ($\epsilon$) as parameters and data go to infinity. We capture this by modeling the percent increase in \textit{reducible} loss at each scale as a quadratic in the hyperparameters:
\begin{equation}\label{eq:static-empirical-model_hessian}
    \mathcal{L}(p, d) \approx \epsilon + \left(\frac{\eta}{d^\kappa} + \frac{\zeta}{p^\iota}\right)\left(1 + L \bX + \left(\bX - \bx_*\right)^T H_{\bx_*} \left(\bX - \bx_*\right)\right)
\end{equation}
Here, $L$ represents residual linear terms necessary to make any quadratic expressible.\footnote{
    One might expect $L = \pmb{0}$ at the optimum. While we do not enforce this assumption in our experiments, we explored it and the conclusions are similar.
} With this model, the scaling law supplies the optimal loss across scales while the quadratic inflates the reducible part according to how far the hyperparameters sit from the optimum.

Since the Hessian is real and symmetric, we can diagonalize it: $H_{\bx_*} = U D U^T$. Let $\Lambda$ be the diagonal matrix of non-zero eigenvalues and take $M_{\bx_*}$ to be the submatrix of $U^T$ that projects onto their corresponding eigenvectors. Letting $\bz_* = M_{\bx_*} \bx_*$ represent the linear subspace of optimal hyperparameters, we can write this model as a quadratic in the (lower-dimensional) projected space:
\begin{equation}\label{eq:static-empirical-model_projection}
    \mathcal{L}(p, d) \approx \epsilon + \left(\frac{\eta}{d^\kappa} + \frac{\zeta}{p^\iota}\right)\left(1 + L \bX + \left(M_{\bx_*} \bX - \bz_*\right)^T \Lambda \left(M_{\bx_*} \bX - \bz_*\right)\right)
\end{equation}
This form enables us to fit the model with a low rank constraint.

We explored several variants of this model, in all cases restricting the Hessian's rank to at most $\gamma = 3$. The \texttt{static} model fixes both the optimum and the curvature across scales as in \Cref{eq:static-empirical-model_projection}. The \texttt{dynamic optima} model keeps the curvature fixed but makes the (latent) optimal hyperparameters a linear function of the log scale: $\bz_* = \bz_0 + \log(p) \bz_{p} + \log(d) \bz_d$. The \texttt{dynamic sensitivity} model instead fixes the optimum and lets the curvature vary, making the eigenvalues of the Hessian power laws of the scale: $\lambda_i p^{\nu_i} d^{\xi_i}$. The \texttt{dynamic} model varies both. For comparison, we include a \texttt{polynomial} model which predicts the percent increase in excess loss using a quadratic polynomial (with no constraint on the rank). We also compare against a \texttt{multilayer perceptron} model that abandons interpretability altogether and predicts percent increase in excess loss as a nonlinear function (neural network) of the hyperparameters and scale.

\begin{table}
    \centering
    \begin{tabular}{lccc}
        \toprule
                                   &  \multicolumn{2}{c}{Validation (MSE)} & \multicolumn{1}{c}{Test (MSE)} \\
                                   &               67M &              134M &                          268M  \\
        \midrule
        Scaling Law                &         0.000011  &         0.000025  &                      0.000020  \\
        \midrule
        Parameters and Tokens Only &         0.011471  &         0.014014  &                      0.019795  \\
        Polynomial                 &         0.004452  &         0.005217  &                      0.008790  \\
        Static                     &         0.004820  &         0.005900  &                      0.010027  \\
        Dynamic Optima             &   \emph{0.002824} &   \emph{0.004401} &                \emph{0.006630} \\
        Dynamic Sensitivity        &         0.005413  &         0.007185  &                      0.010813  \\
        Dynamic                    &         0.003672  &         0.006806  &                      0.010102  \\
        Multilayer Perceptron      &         0.008866  &         0.012003  &                      0.015464  \\
        \bottomrule
    \end{tabular}
    \caption{
        \emph{The mean-squared error (MSE) of different models for the hyperparameter loss surface extrapolating from small to large scales using our model ladder and decaying the learning rate.} The scaling law predicts loss at the optimal hyperparameters; the rest predict all configurations in the asymptotic regime. Each column's best loss is italicized (excluding the scaling law). The static quadratic that ignores scale is already a strong baseline; varying the optimum appears to help while varying the curvature seems to hurt. The \texttt{dynamic optima} model performs best here; however, the models' ranking is sensitive to optimization details like the initialization. Every variant improves over predicting from parameters and tokens alone (ignoring the hyperparameters), but none reaches the error attained by the scaling law when predicting the loss at the optimal hyperparameters; in other words, no model fully captures the surface.
    }
    \label{tab:aggregated-mse_ladder-decay}
\end{table}

\Cref{tab:aggregated-mse_ladder-decay} compares these variants, fit on the small scales and evaluated on the larger held-out ones via mean-squared error (MSE). Ignoring the scale's effect on the hyperparameters (\texttt{static}) already performs surprisingly well. The dynamic variants extrapolate with mixed success: varying the optimum appears to help, while varying the curvature seems to hurt. The nonlinear model is the most flexible; here it performs the worst of all models that incorporate the hyperparameters, though it can also perform the best in similar setups depending on optimization details like the initialization. While every variant beats the baseline of ignoring the hyperparameters and predicting from the parameters and tokens alone, none fully captures the loss surface. If one did, it would approach the much smaller error achieved by the scaling law when predicting the optimal hyperparameters.

%% file: contents/appendix/additional_theoretical_results.tex
\section{Additional Theoretical Results}
\label{app:additional-theoretical-results}

\subsection{Scaling law error propagation}
\label{sec:scaling-law-error-propagation}

Scaling laws are fitted to losses measured at a handful of scales and then extrapolated to a larger ones. Each measured loss, however, carries error from a finite tuning budget (\cref{app:background:how-close-does-random-search-get}), random seeds, or estimation noise. If the irreducible error were known, this problem would reduce to fitting a power law: essentially a linear regression in log-log space. In practice, the irreducible error is not known, but the fitting procedure can behave quite similarly. We ask a simple question: if the loss at one scale is off by $\epsilon$, how much does that corrupt the loss we \emph{extrapolate} to at a larger scale? For the linear case, the answer is exact, closed form and can be retrieved from standard literature on linear regression.

Throughout, we index the $K$ \emph{scales} by $j = 1, \dots, K$; each scale contributes one data point to the regression. We reserve $n$ (and $M$) for the number of \emph{samples} used to \emph{measure} the loss at a given scale, for instance, the best-of-$M$ score $T_M$ from a tuning run (\cref{app:background:how-close-does-random-search-get}), so that the measurement error $\epsilon$ at a scale can itself be understood through the sampling analysis of the previous section. \Cref{fig:scaling-error-propagation} illustrates the whole construction.

\paragraph{Scaling law fit as log-log regression.}
Assume a power-law scaling law, $L = A\,x^{-b}$, where $x$ is the scale (compute, parameters, or data) and $L$ the loss. Taking logarithms linearizes it,
\begin{equation}\label{eq:loglog}
    y = \beta_0 + \beta_1 u, \qquad u := \log x, \quad y := \log L,
\end{equation}
with $\beta_0 = \log A$ and slope $\beta_1 = -b$. Given measurements $\{(u_j, y_j)\}_{j=1}^{K}$ at the $K$ scales, we estimate $(\beta_0, \beta_1)$ by ordinary least squares \citep{brook2018applied, montgomery2022introduction}.
%\footnote{   We fit the pure power law; an additive irreducible term ($L = E + A x^{-b}$) breaks the exact log-linearity and is fit by nonlinear least squares. The linear analysis here is the leading-order behavior when the power-law term dominates.}

\paragraph{Extrapolation as a weighted sum of the data.}
Ordinary least squares is linear in the responses, so both the fitted coefficients and any prediction are weighted sums of the observed $y_j$ \citep{brook2018applied, montgomery2022introduction}. Writing $\bar u = \tfrac1K \sum_k u_k$ and $S_{uu} = \sum_k (u_k - \bar u)^2$, the slope estimate is
\begin{equation}\label{eq:ols-slope}
    \hat\beta_1 = \sum_{j=1}^{K} w_j\, y_j, \qquad w_j = \frac{u_j - \bar u}{S_{uu}},
\end{equation}
and the intercept is $\hat\beta_0 = \bar y - \hat\beta_1 \bar u$. Substituting into \Cref{eq:loglog}, the predicted log-loss at any scale $u_\star$ --- in particular one we extrapolate to --- is itself a weighted sum,
\begin{equation}\label{eq:prediction-weights}
    \hat y(u_\star) = \sum_{j=1}^{K} h_j(u_\star)\, y_j, \qquad h_j(u_\star) = \frac{1}{K} + \frac{(u_\star - \bar u)(u_j - \bar u)}{S_{uu}}.
\end{equation}

\begin{wrapfigure}{r}{0.45\textwidth}
    \vspace{-0.9cm}
    \centering
    \definecolor{plumdark}{HTML}{4A2472}
    \definecolor{plummed}{HTML}{7B5EA7}
    \definecolor{plumlight}{HTML}{C4A0C8}
    \begin{tikzpicture}
    \begin{axis}[
        clip=false, width=0.5\textwidth, height=0.38\textwidth,
        xmin=0.3, xmax=4.7, ymin=-1.15, ymax=1.55,
        axis x line=middle, axis y line=left,
        xtick={1,2,3,4}, xticklabels={$u_1$,$u_2$,$u_3$,$u_4$},
        ytick={-1,0,1}, tick label style={font=\scriptsize},
        ylabel={leverage $h_j(u_\star)$}, ylabel style={font=\scriptsize},
        x axis line style={-},
    ]
      \addplot[plummed, dashed, thick, domain=0.55:4.45, forget plot] {0.25+0.7*(x-2.5)};
      \addplot[ycomb, plumdark, very thick, mark=*, mark size=2.5pt]
          coordinates {(1,-0.80) (2,-0.10) (3,0.60) (4,1.30)};
      \draw[gray, dashed, thin] (axis cs:2.5,-1.15) -- (axis cs:2.5,0.25);
      \node[gray, font=\scriptsize, anchor=north] at (axis cs:2.5,-0.02) {$\bar u$};
      \node[plumdark, font=\scriptsize, anchor=west, align=left] at (axis cs:1.05,1.25)
          {$|h_j|$ grows linearly\\ with $|u_j-\bar u|$};
    \end{axis}
    \end{tikzpicture}
    \caption{
        Leverage of each measurement. The leverage $h_j(u_\star)$ (\Cref{eq:prediction-weights}) measures how strongly each measurement steers the extrapolated prediction. Its magnitude grows linearly with the distance $|u_j - \bar u|$ from the centroid, and its sign flips across $\bar u$, so the endpoints carry the most weight.
    }
    \label{fig:leverage}
    \vspace{-0.6cm}
\end{wrapfigure}

The weight $h_j(u_\star)$ is the sensitivity of the extrapolated prediction to the $j$-th measurement, we will refer to it as the \emph{leverage} of point $j$ \citep{hastie2009elements}. $|h_j(u_\star)|$ is large when $j$ is far from the center.
We illustrate this in \Cref{fig:leverage}.
Note that, because $u_\star$ lies outside the data range $|u_\star - \bar u|$ is large which is why extrapolated predictions are so sensitive to the measurements with the highest leverage.

\paragraph{Propagating errors.}
Suppose the measurement at scale $j$ is perturbed, $y_j \mapsto y_j + \epsilon$, while the others are exact. Because \Cref{eq:prediction-weights} is linear, the prediction shifts by exactly
\begin{equation}\label{eq:single-error}
    \delta \hat y(u_\star) = h_j(u_\star)\, \epsilon.
\end{equation}
As is standard for scaling laws, take the scales equally spaced in log space, $u_j = u_1 + (j - 1)\Delta$. Then $u_j - \bar u = (j - \bar\jmath)\Delta$ with center $\bar\jmath = \tfrac{K+1}{2}$, and $S_{uu} = \Delta^2 \sum_k (k - \bar\jmath)^2 = \Delta^2\, \tfrac{K(K^2 - 1)}{12}$. Extrapolating from the corrupted scale $j$ to a scale $i$ steps further, $u_\star = u_{j+i}$, the spacing $\Delta$ cancels and \Cref{eq:single-error} becomes
\begin{equation}\label{eq:propagation}
    \delta \hat y_{j+i} = \epsilon \left[\, \frac{1}{K} + \frac{12\, (j - \bar\jmath)(j + i - \bar\jmath)}{K(K^2 - 1)} \,\right].
\end{equation}
Note that, an error $\epsilon$ at scale $j$ shifts the extrapolated log-loss at scale $j + i$ by a factor that depends on how many scales $j$ is from the center $\bar\jmath$ and how far the extrapolation $i$ is. Errors at the endpoints of the fitting range are the most damaging, while an error at the center ($j = \bar\jmath$) contributes only the $\nicefrac{1}{K}$ intercept shift and is not amplified by extrapolation.

\begin{figure}[!hbt]
    \centering
    \definecolor{plumdark}{HTML}{4A2472}
    \definecolor{plummed}{HTML}{7B5EA7}
    \definecolor{plumlight}{HTML}{C4A0C8}
    \begin{tikzpicture}[>=stealth]
    \begin{axis}[
        name=bax,
        clip=false, scale only axis, width=0.8\textwidth, height=0.42\textwidth,
        xmin=-0.4, xmax=6.9, ymin=1.4, ymax=5.7,
        axis lines=left,
        xtick={1,2,3,4,6}, ytick={2,3,4,5},
        xticklabel=\empty, yticklabel=\empty,
        xmajorgrids=true, ymajorgrids=true,
        major grid style={draw=gray!20, very thin},
        major tick length=0pt,
        every axis plot/.append style={very thick},
        legend style={font=\scriptsize, draw=none, fill=none,
            at={(0.5,-0.62)}, anchor=north, legend columns=1, row sep=1pt},
        legend cell align=left,
    ]
      \addplot[draw=none, fill=plumlight, fill opacity=0.18, forget plot]
          coordinates {(4,1.4) (4,5.7) (6.9,5.7) (6.9,1.4)} \closedcycle;
      \node[font=\scriptsize, gray, anchor=north] at (axis cs:5.4,5.68) {extrapolation};
      \addplot[plumdark, domain=0:6.5] {5-0.5*x};
      \addlegendentry{true scaling law}
      \addplot[plumlight, dotted, line width=1.4pt, domain=0:6.5] {5.06-0.53*x};
      \addlegendentry{estimate with error at small-leverage scale, good fit}
      \addplot[plummed, dashed, domain=0:6.5] {4.7-0.32*x};
      \addlegendentry{estimate with error at large-leverage scale, bad fit}
      \addplot[only marks, mark=*, plumdark, mark size=2.2pt, forget plot]
          coordinates {(1,4.5) (2,4) (3,3.5) (4,3)};
      \addplot[only marks, mark=o, plummed, mark size=3pt, forget plot] coordinates {(4,3.6)};
      \draw[->, plummed, semithick, shorten <=1pt, shorten >=1pt] (axis cs:4,3.05) -- (axis cs:4,3.55);
      \node[plummed, font=\scriptsize, anchor=south west] at (axis cs:4.05,3.5) {$\epsilon$ far from center $u_4$};
      \draw[<->, plummed, semithick] (axis cs:6.15,2.02) -- (axis cs:6.15,2.78);
      \node[plummed, font=\scriptsize, anchor=west] at (axis cs:6.25,2.42)
          {$\delta\hat y = h_4\epsilon$};
      \addplot[only marks, mark=o, plumlight, mark size=3pt, forget plot] coordinates {(2,4.6)};
      \draw[->, plumlight, semithick, shorten <=1pt, shorten >=1pt] (axis cs:2,4.05) -- (axis cs:2,4.55);
      \node[font=\scriptsize, anchor=south, text=plumlight!80!black] at (axis cs:2,4.62)
          {$\epsilon$ near center $u_2$};
      \draw[<->, plumlight, semithick] (axis cs:5.85,1.9) -- (axis cs:5.85,2.02);
      \node[font=\scriptsize, anchor=east, text=plumlight!70!black] at (axis cs:5.75,1.75)
          {$\delta\hat y = h_2\epsilon$};
      \draw[gray, dashed, thin] (axis cs:2.5,1.4) -- (axis cs:2.5,3.75);
      \draw[gray, dashed, thin] (axis cs:6,1.4) -- (axis cs:6,3.0);
      \node[font=\scriptsize, anchor=north] at (axis cs:1,1.38) {$u_1$};
      \node[font=\scriptsize, anchor=north] at (axis cs:2,1.38) {$u_2$};
      \node[font=\scriptsize, anchor=north] at (axis cs:3,1.38) {$u_3$};
      \node[font=\scriptsize, anchor=north] at (axis cs:4,1.38) {$u_4$};
      \node[font=\scriptsize, anchor=north] at (axis cs:2.5,1.20) {$\bar u$};
      \node[font=\scriptsize, anchor=north] at (axis cs:6,1.38) {$u_\star$};
      \node[font=\small, anchor=west] at (axis cs:7.0,1.4) {$u=\log x$};
      \node[font=\small, anchor=south, rotate=90] at (axis cs:-0.4,5.2) {$y=\log L$};
    \end{axis}
    % second x-axis (below the first): centered index
    \begin{axis}[
        at={(bax.south west)}, anchor=north west, yshift=-22pt,
        clip=false, scale only axis, width=0.8\textwidth, height=1pt,
        xmin=-0.4, xmax=6.9, ymin=0, ymax=1,
        axis x line=bottom, hide y axis,
        xtick={1,2,3,4},
        xticklabels={$-\tfrac{3}{2}$,$-\tfrac{1}{2}$,$+\tfrac{1}{2}$,$+\tfrac{3}{2}$},
        x tick label style={font=\scriptsize, anchor=north},
    ]
      \node[font=\small, anchor=west] at (axis cs:7.0,0) {$j-\hat\jmath$};
    \end{axis}
    % third x-axis (below the second): leverage
    \begin{axis}[
        name=levax,
        at={(bax.south west)}, anchor=north west, yshift=-44pt,
        clip=false, scale only axis, width=0.8\textwidth, height=1pt,
        xmin=-0.4, xmax=6.9, ymin=0, ymax=1,
        axis x line=bottom, hide y axis,
        xtick={1,2,3,4},
        xticklabels={$-0.8$,$-0.1$,$+0.6$,$+1.3$},
        x tick label style={font=\scriptsize, anchor=north},
    ]
      \node[font=\small, anchor=west] at (axis cs:7.0,0) {$h_j(u_\star)$};
    \end{axis}
    % extend the left (y) axis down so it meets the bottom-most x-axis
    \draw[thin] (bax.south west) -- (levax.south west);
    \end{tikzpicture}
    \caption{
        Consequence for extrapolation. Losses are measured at $K$ equally spaced scales $u_1, \dots, u_K$ (centroid $\bar u$) and fit by ordinary least squares (solid; \Cref{eq:loglog} and \ref{eq:ols-slope}). Perturbing a measurement by the same $\epsilon$ shifts the extrapolated prediction at $u_\star$ by $\delta\hat y = h_j(u_\star)\,\epsilon$ (\Cref{eq:single-error}): a high-leverage point far from the center ($u_4$, dashed) \emph{amplifies} the error, while a low-leverage point near the center ($u_2$, dotted) barely moves it. Because $u_\star$ lies far outside the data range (shaded), it amplifies the effect of the leverage further ($h_4(u_\star) > 1$ here). The lower axes read each measurement's centered index $j-\hat\jmath$ and its leverage $h_j(u_\star)$.
    }
    \label{fig:scaling-error-propagation}
\end{figure}

\subsection{How many hyperparameter runs are needed?}
\label{app:background:how-close-does-random-search-get}

How many hyperparameter runs are needed to approximate the optimal perplexity at a given scale? Or in other words, how quickly does the sample minimum of the noisy quadratic converge to the best attainable performance $\alpha$? We will see that the answer depends on the intrinsic dimension $\gamma$ of the hyperparameter space.

Given the noisy quadratic limit (see \Cref{app:background:the-noisy-quadratic-limit}), whose notation ($\alpha$, $\beta$, $\gamma$, $\sigma$, and the CDF $F$) we adopt throughout, we can quantify how close random search gets to the best attainable performance $\alpha$ after $M$ evaluations, i.e. how many hyperparameter runs get us a good estimate for the best attainable performance at a given scale. Let $Y_1, \dots, Y_M$ be i.i.d.\ scores drawn within the asymptotic regime so each $Y_i$ follows the noisy quadratic distribution of \Cref{eq:noisy-quadratic-limit}, and let $T_M = \min_{i \leq M} Y_i$ be the best score found.

First, we find the CDF of $T_M$, by reasoning about the complement: the best score exceeds a threshold $y$ only when \emph{every} draw does, and a single draw exceeds $y$ with probability $\Prob(Y_i > y) = 1 - F(y)$, and because we draw independent samples,
\begin{equation}\label{eq:min-survival}
    \Prob(T_M > y) = \prod_{i=1}^{M} \Prob(Y_i > y) = \prod_{i=1}^{M} \underbrace{\bigl(1 - F(y)\bigr)}_{=\,\Prob(Y_i > y)} = \bigl(1 - F(y)\bigr)^M.
\end{equation}
Taking the complement gives the CDF of the minimum $T_M$,
\begin{equation}\label{eq:min-cdf}
    \Prob(T_M \leq y) = 1 - \bigl(1 - F(y)\bigr)^M.
\end{equation}

Second, we determine how $T_M$ approaches the optimum $\alpha$ as $M$ grows by consulting Extreme Value Theory (EVT)\citep{gnedenko1943distribution, de2006extreme}
\footnote{
    Extreme-value theory is usually stated for maxima; the result for minima follows by applying it to $-Y$.
}.
We will see there are two regimes: the noiseless case ($\sigma = 0$) and the noisy case ($\sigma > 0$) for which we will come to different conclusions.
Consider the noiseless case ($\sigma = 0$), where $\alpha$ is the lower endpoint of the support and the tail is the power law of \Cref{eq:quadratic-cdf}. A distribution with a finite endpoint and a polynomially decaying tail lies in the \emph{Weibull domain of attraction} of EVT: its sample minima, once recentered at the endpoint and rescaled, converge to a Weibull law. Concretely, take the natural scale $a_M := (\beta - \alpha)\, M^{-2/\gamma}$, defined implicitly by $F(\alpha + a_M) = 1/M$. Substituting $u = a_M t$ yields $\bigl(1 - F(\alpha + a_M t)\bigr)^M \to \exp(-t^{\gamma/2})$, so the rescaled gap converges in distribution to 

\begin{equation}\label{eq:sample-min-weibull}
    \frac{T_M - \alpha}{a_M} \;\xrightarrow{\;d\;}\; \operatorname{Weibull}(k), \qquad k = \tfrac{\gamma}{2},
\end{equation}
where $\operatorname{Weibull}(k)$ is the Weibull distribution of shape $k$ and unit scale, see \Cref{fig:weibull-shapes} for an illustration.
\begin{figure}[!hbt]
    \centering
    \definecolor{plumdark}{HTML}{4A2472}
    \definecolor{plummed}{HTML}{7B5EA7}
    \definecolor{plumlight}{HTML}{C4A0C8}
    \begin{tikzpicture}
        \begin{axis}[
            clip=false,
            width=0.6\textwidth, height=0.36\textwidth,
            domain=0.05:4, samples=200,
            ymin=0, ymax=1.5, xmin=0,
            axis lines=left,
            xlabel={rescaled gap $x = (T_M - \alpha)/a_M$},
            ylabel={density},
            xlabel style={font=\scriptsize}, ylabel style={font=\scriptsize},
            tick label style={font=\scriptsize},
            legend style={font=\scriptsize, draw=none, fill=none, at={(0.03,0.97)}, anchor=north west},
            every axis plot/.append style={very thick},
        ]
            \addplot[draw=none, fill=plummed, fill opacity=0.13, forget plot]
                coordinates {(2.5,0) (2.5,1.5) (4,1.5) (4,0)} \closedcycle;
            \addplot[plumlight, domain=0.08:4] {0.5*x^(-0.5)*exp(-(x^0.5))};
            \addlegendentry{$\gamma=1\ (k=\tfrac12)$}
            \addplot[plummed]   {exp(-x)};
            \addlegendentry{$\gamma=2\ (k=1)$}
            \addplot[plumdark]  {1.5*(x^0.5)*exp(-(x^1.5))};
            \addlegendentry{$\gamma=3\ (k=\tfrac32)$}
            \node[font=\scriptsize, plumdark, align=center] at (axis cs:3.25,0.75)
                {upper tail\\[-1pt] $\lesssim e^{-x^{k}}$};
            \addplot[only marks, mark=*, plumdark, mark size=2pt, forget plot]
                coordinates {(0,0)};
            \draw[->, plumdark, thin] (axis cs:-0.55,0.3) -- (axis cs:-0.02,0.02);
            \node[font=\scriptsize, plumdark, anchor=east] at (axis cs:-0.6,0.34)
                {endpoint $\alpha$};
        \end{axis}
    \end{tikzpicture}
    \caption{
        Weibull($k$) densities $f(x) = k\, x^{k-1} e^{-x^{k}}$ (and CDF $F(x) = 1 - e^{-x^{k}}$).  The shape $k = \gamma/2$ governs how quickly probability accumulates away from the endpoint $\alpha$: a small effective dimension ($\gamma = 1$) piles mass near the optimum, while larger $\gamma$ pushes the typical gap outward. In the shaded \emph{upper tail} every density decays at least as fast as $e^{-x^{k}}$, so a vanishing fraction of both the mass and the mean sits out there, which illustrates why this function is uniformly integrable, meaning that the mean exists.
    }
    \label{fig:weibull-shapes}
\end{figure}

Third, as the limit in \Cref{eq:sample-min-weibull} is a convergence in distribution, it does not by itself imply convergence of its mean $E[(T_M-\alpha)/a_M]$. For this, we have to additionally show that the rescaled minimum is \emph{uniformly integrable}, which ensures that the mean exists. To see this condition is fulfilled, note that the upper tail is bounded by $\exp(-t^{\gamma/2})$.
Thus, the rescaled minimum is uniformly integrable and the mean converges to that of the limiting Weibull distribution;
\begin{equation}\label{eq:sample-min-rescaled-mean}
    \E\!\left[\frac{T_M - \alpha}{a_M}\right] \;\longrightarrow\; \E\bigl[\operatorname{Weibull}(k)\bigr] = \Gamma\!\left(1 + \tfrac{1}{k}\right),
\end{equation}
 where $\Gamma$ is the gamma function \citep{johnson1994continuous}.

Finally, we have all the pieces to show how quickly the gap between the observed and optimal minima, $\E[T_M] - \alpha$, closes as the number of samples $M$ increases. Its dependence on $M$ follows from \Cref{eq:sample-min-weibull} by multiplying through by the scale $a_M = (\beta - \alpha)\, M^{-2/\gamma}$ and setting $k = \gamma/2$,
\begin{equation}\label{eq:sample-min-mean}
    \E[T_M] - \alpha \;\sim\;
    a_M \, \Gamma\!\left(1 + \tfrac{1}{k}\right) \;=\; \underbrace{(\beta - \alpha)\, \Gamma\!\left(1 + \tfrac{2}{\gamma}\right)}_{=: C(\gamma)} \; M^{-2/\gamma}.
\end{equation}

Here and below, $\sim$ denotes \emph{asymptotic equivalence}: $f(M) \sim g(M)$ means $f(M)/g(M) \to 1$ as $M \to \infty$, not proportionality or ``distributed as''. For the practically relevant cases $\gamma \in \{1, 2\}$ the mean can even be evaluated exactly,\footnote{
    The ``$\sim$'' in \Cref{eq:sample-min-mean} is asymptotic: it is the $M \to \infty$ limit of the Weibull law \Cref{eq:sample-min-weibull}, whose mean is the standard $\Gamma(1 + 1/k)$. For the noiseless quadratic we can go further and evaluate the mean exactly at finite $M$. Using the tail identity $\E[Z] = \int_0^\infty \Prob(Z > u)\, \d u$ for the nonnegative $Z = T_M - \alpha$ together with the survival function \Cref{eq:min-survival}, the gap is a Beta integral,
    \begin{equation}\label{eq:sample-min-exact}
        \E[T_M] - \alpha
        = \int_0^{\beta - \alpha} \bigl(1 - F(\alpha + u)\bigr)^M\, \d u
        = (\beta - \alpha)\, \Gamma\!\left(1 + \tfrac{2}{\gamma}\right) \frac{\Gamma(M + 1)}{\Gamma\!\left(M + 1 + \nicefrac{2}{\gamma}\right)}.
    \end{equation}
    The ratio of the two gamma functions decays like $M^{-2/\gamma}$ as $M \to \infty$, recovering \Cref{eq:sample-min-mean}. Since the number of runs $M$ is an integer, $\Gamma(M + 1) = M!$, and whenever $2/\gamma$ is a positive integer the ratio collapses to a rational function of $M$, giving the exact expressions in \Cref{eq:sample-min-mean-exact}.
}
giving the main result of this section:
%
%\begin{tcolorbox}[colback=plumbox, colframe=plumdark, boxrule=0.8pt, arc=2pt,
    left=8pt, right=8pt, top=4pt, bottom=4pt]
\begin{equation}\label{eq:sample-min-mean-exact}
    \E[T_M] - \alpha \;
    \begin{cases}
         \sim\; C(\gamma)\, M^{-2/\gamma}, & \text{for all } \gamma,\\[1.6ex]
        = \dfrac{2(\beta - \alpha)}{(M+1)(M+2)} \;\sim\; 2(\beta - \alpha)\, M^{-2}, & \text{for } \gamma = 1,\\[1.6ex]
        = \dfrac{\beta - \alpha}{M+1} \;\sim\; (\beta - \alpha)\, M^{-1}, & \text{for } \gamma = 2.
    \end{cases}
\end{equation}
%\end{tcolorbox}

Thus the best score approaches $\alpha$ as a power law in the number of evaluations, $M^{-2/\gamma}$, with the exponent set by the effective number of hyperparameters. Each additional effective hyperparameter flattens the convergence exponent.

As a concrete example, say we ran 100 hyperparameter evaluations at a scale where $\gamma = 2$. Assuming all other scales have the same $\alpha$ and $\beta$, we can compare how many runs would be needed at other scales where $\gamma = 1, 3$ to reach the same quality, i.e.\ the same gap.
Matching the gaps in \Cref{eq:sample-min-mean-exact}, i.e.\ $C(\gamma)\, M_\gamma^{-2/\gamma} = C(2)\, 100^{-1}$, gives $M_\gamma = \bigl[\,100\, C(\gamma)/C(2)\,\bigr]^{\gamma/2}$. Since $C(1) = 2(\beta - \alpha)$, $C(2) = (\beta - \alpha)$, and $C(3) \approx 0.903\,(\beta - \alpha)$, the constant ratios are $C(1)/C(2) = 2$ and $C(3)/C(2) \approx 0.903$, so
\begin{equation*}
    M_1 = (100 \cdot 2)^{1/2}  \approx 14
    \quad\text{and}\quad
    M_3 = (100 \cdot 0.903)^{3/2}  \approx 858.
\end{equation*}
So dropping from two effective hyperparameters to one cuts the tuning cost from $100$ runs to about $14$ (a $7\times$ saving), while adding one raises it to nearly $900$ (a further $8.6\times$).

Note that we assumed $\sigma = 0$ for our derivation, when $\sigma > 0$, the unbounded Gaussian tail means we can no longer rely on the same analysis as the EVT is now in the heavy tail regime.

\subsection{Compute-optimal tuning across scales}
\label{sec:compute-optimal-tuning}

In this section we will combine the results from the previous two sections to answer the question: Given a fixed compute budget, how should we split it across scales to minimize the error at the scale we extrapolate to?

For our analysis, we make the following assumptions, 
\begin{enumerate}[noitemsep]
    \item \textbf{Tuning error per scale.} At scale $j$ we run random search with $M_j$ configurations and record the best. Its expected gap to the optimum is $\epsilon_j = A_j\, M_j^{-2/\gamma_j}$ with $A_j = (\beta_j - \alpha_j)\,\Gamma(1 + 2/\gamma_j)$ (\Cref{eq:sample-min-mean}), where $\gamma_j$ is the effective number of hyperparameters at that scale.
    \item \textbf{Cost.} Compute per run scales linearly with the parameter count, so one run at scale $j$ costs $\propto N_j$. The scales are geometric, $N_j = N_1 \rho^{\,j-1}$ (equidistant in $\log N$), and the total budget is $C = \sum_{j=1}^{K} M_j N_j$.
    \item \textbf{Propagation.} An error $\epsilon_j$ shifts the extrapolated prediction by $h_j\,\epsilon_j$ (\Cref{eq:single-error}), where $h_j = h_j(u_\star)$ is the leverage of scale $j$ at the target $u_\star$. We minimize the propagated-error bound $E_\star = \sum_{j} |h_j|\, \epsilon_j$.\footnote{
        Using the triangle inequality on $\sum_j h_j \epsilon_j$ gives a sign-robust objective. Minimizing the variance proxy $\sum_j h_j^2 \epsilon_j^2$ instead replaces the exponent $\nicefrac{\gamma_j}{(\gamma_j + 2)}$ below by $\nicefrac{\gamma_j}{(\gamma_j + 4)}$ and leaves the qualitative allocation unchanged.
    }
    \item \textbf{Effective dimension.} Empirically $\gamma_j \in \{1, 2\}$, and it \emph{decreases} with scale: the low-compute scales have $\gamma_j = 2$, and beyond some scale the effective dimension drops to $\gamma_j = 1$.
\end{enumerate}

\paragraph{The optimization problem.}
Substituting the tuning error $\epsilon_j = A_j M_j^{-2/\gamma_j}$ into the propagated-error bound, the quantity we want to minimize with respect to the number of runs $M_j$ is
\begin{equation}\label{eq:objective}
    E_\star(M_1, \dots, M_K) = \sum_{j=1}^{K} |h_j|\, \epsilon_j = \sum_{j=1}^{K} |h_j|\, A_j\, M_j^{-2/\gamma_j},
\end{equation}
however this error minimization is subject to the compute budget constraint \Cref{eq:constraint};
\begin{equation}\label{eq:constraint}
    \sum_{j=1}^{K} M_j N_j = C.
\end{equation}
We attach the constraint with a Lagrangian multiplier $\lambda$ and minimize the loss
\begin{equation}\label{eq:lagrangian}
    \mathcal{L}(M_1, \dots, M_K, \lambda) = \sum_{j=1}^{K} |h_j|\, A_j\, M_j^{-2/\gamma_j} + \lambda\left(\sum_{j=1}^{K} M_j N_j - C\right).
\end{equation}
Differentiating with respect to a single $M_j$ gives ,
\begin{equation}\label{eq:foc}
    \frac{\partial \mathcal{L}}{\partial M_j}
    = -\frac{2}{\gamma_j}\, |h_j|\, A_j\, M_j^{-2/\gamma_j - 1} + \lambda N_j
    = 0.
\end{equation}
Rearranged for $\lambda$;
\begin{equation}\label{eq:equimarginal}
    \underbrace{\frac{1}{N_j}\cdot\frac{2}{\gamma_j}\, |h_j|\, A_j\, M_j^{-2/\gamma_j - 1}}_{\text{error reduction per unit compute at scale } j} = \lambda \quad\text{for every active scale } j.
\end{equation}
Solve for $M_j$ in \Cref{eq:foc},
\begin{equation}\label{eq:solve-step}
    M_j^{2/\gamma_j + 1} = \frac{2\, |h_j|\, A_j}{\gamma_j\, \lambda\, N_j}.
\end{equation}
The exponent on the left combines as $\dfrac{2}{\gamma_j} + 1 = \dfrac{\gamma_j + 2}{\gamma_j}$, so raising both sides to the reciprocal power $\dfrac{\gamma_j}{\gamma_j + 2}$ gives the optimal number of runs,
\begin{equation}\label{eq:opt-runs}
    M_j^\star = \left(\frac{2\,|h_j|\, A_j}{\gamma_j\, \lambda\, N_j}\right)^{\frac{\gamma_j}{\gamma_j + 2}}
    \;\propto\; \left(\frac{|h_j|}{\gamma_j\, N_j}\right)^{\frac{\gamma_j}{\gamma_j + 2}}.
\end{equation}
\paragraph{Example.}
To make the allocation concrete, take $K = 4$ equally spaced scales and extrapolate one step beyond the fitted range, $u_\star = u_{K+1}$ (i.e.\ $i = 1$ in \Cref{eq:propagation}). With centre $\bar\jmath = \tfrac{K+1}{2} = \tfrac52$, the target sits at $u_\star - \bar u = \bar\jmath\,\Delta$ and the index form of the leverage collapses to a single line,
\begin{equation}\label{eq:example-leverage}
    h_j = \frac{1}{K} + \frac{12\,\bar\jmath\,(j - \bar\jmath)}{K(K^2 - 1)}
        = \frac{j - 2}{2},
    \qquad\text{so}\qquad
    \bigl(|h_1|, |h_2|, |h_3|, |h_4|\bigr) = \left(\tfrac12,\; 0,\; \tfrac12,\; 1\right).
\end{equation}
The endpoint nearest the target ($j = 4$) carries the most leverage, while the near-central scale ($j = 2$) carries essentially none. Following the fourth assumption, the effective dimension drops across the centroid: $\gamma_j = 2$ for the small scales below the centre ($j = 1, 2$) and $\gamma_j = 1$ for the large scales above it ($j = 3, 4$). The per-run costs are geometric in the index, $N_j = N_1\, \rho^{\,j-1}$, i.e.\ linear in log space, $\log N_j = \log N_1 + (j-1)\log\rho$.

Substituting into \Cref{eq:opt-runs} and reading off the exponent $\tfrac{\gamma_j}{\gamma_j + 2}$ ($= \tfrac12$ where $\gamma_j = 2$, $= \tfrac13$ where $\gamma_j = 1$), the optimal number of runs at each scale is
\begin{equation}\label{eq:example-runs}
    M_j^\star \;\propto\; \left(\frac{|h_j|}{\gamma_j\, N_j}\right)^{\frac{\gamma_j}{\gamma_j + 2}}
    = \begin{cases}
        \bigl(\,|h_j| / (2\, N_1 \rho^{\,j-1})\,\bigr)^{1/2}, & j \le 2 \quad (\gamma_j = 2),\\[6pt]
        \bigl(\,|h_j| / (N_1 \rho^{\,j-1})\,\bigr)^{1/3}, & j \ge 3 \quad (\gamma_j = 1).
    \end{cases}
\end{equation}
Multiplying by $N_j$ gives the compute placed at each scale,
\begin{equation}\label{eq:opt-compute}
    C_j^\star = N_j\, M_j^\star \;\propto\; \left(\frac{|h_j|}{\gamma_j}\right)^{\frac{\gamma_j}{\gamma_j + 2}} N_j^{\frac{2}{\gamma_j + 2}},
\end{equation}
whose exponent on $N_j$ ($\tfrac12$ then $\tfrac23$) is below one, so compute grows \emph{sublinearly} with a scale's per-run cost. Within the large-scale regime the two upper scales are tied together cleanly,
\begin{equation}\label{eq:example-ratio}
    \frac{M_4^\star}{M_3^\star} = \left(\frac{|h_4|}{|h_3|}\cdot\frac{N_3}{N_4}\right)^{1/3} = \left(\frac{2}{\rho}\right)^{1/3},
    \qquad
    \frac{C_4^\star}{C_3^\star} = 2^{1/3}\,\rho^{2/3}.
\end{equation}
For any $\rho > 2$ the larger scale receives \emph{fewer} runs yet \emph{more} compute than its neighbour: a handful of expensive runs at the top outweighs many cheap ones below. \Cref{fig:compute-allocation} plots the allocation for $\rho = 4$ ($N_j = 1, 4, 16, 64$): the number of runs is largest at the cheap bottom scale and tapers off, while the compute share climbs monotonically and piles up at the top scale. The near-central scale $j = 2$, with $h_2 \approx 0$, drops out entirely and is kept only by its single anchoring run.

\begin{figure}[!htb]
    \centering
    \definecolor{plumdark}{HTML}{4A2472}
    \definecolor{plummed}{HTML}{7B5EA7}
    \definecolor{plumlight}{HTML}{C4A0C8}
    \begin{tikzpicture}
    \begin{axis}[
        name=cax,
        clip=false, scale only axis, width=0.6\textwidth, height=0.4\textwidth,
        xmin=0.4, xmax=4.6, ymin=0, ymax=80,
        axis y line=left, axis x line=bottom,
        xtick={1,2,3,4}, xticklabels={$u_1$,$u_2$,$u_3$,$u_4$},
        ytick={0,20,40,60,80},
        tick label style={font=\scriptsize},
        ylabel={\tikz{\draw[draw=plummed, fill=plumlight, fill opacity=0.6] (0,0) rectangle (6pt,6pt);}\;compute share $C_j^\star/C$ (\%)},
        ylabel style={font=\scriptsize, text=plummed},
        y tick label style={text=plummed},
        major tick length=0pt,
        ymajorgrids=true, major grid style={draw=gray!20, very thin},
    ]
      \addplot[ybar, bar width=15pt, draw=plummed, fill=plumlight, fill opacity=0.6]
          coordinates {(1,8.6) (2,1.1) (3,21.6) (4,68.7)};
      \draw[gray, dashed, thin] (axis cs:2.5,0) -- (axis cs:2.5,80);
      \node[gray, font=\scriptsize, anchor=north] at (axis cs:1.5,79) {$\gamma_j = 2$};
      \node[gray, font=\scriptsize, anchor=north] at (axis cs:3.5,79) {$\gamma_j = 1$};
    \end{axis}
    \begin{axis}[
        at={(cax.south west)}, anchor=south west,
        clip=false, scale only axis, width=0.6\textwidth, height=0.4\textwidth,
        xmin=0.4, xmax=4.6, ymin=0, ymax=36,
        axis y line=right, axis x line=none, xtick=\empty,
        ytick={0,10,20,30},
        tick label style={font=\scriptsize},
        ylabel={\tikz{\node[circle, fill=plumdark, inner sep=0pt, minimum size=5pt]{};}\;runs $M_j^\star$},
        ylabel style={font=\scriptsize, text=plumdark},
        y tick label style={text=plumdark},
        major tick length=0pt,
    ]
      \addplot[plumdark, very thick, mark=*, mark size=2.4pt]
          coordinates {(1,32) (2,1) (3,5) (4,4)};
    \end{axis}
    \end{tikzpicture}
    \caption{
        Compute-optimal allocation across $K = 4$ scales for the example of \Cref{eq:example-leverage} and \ref{eq:example-runs} ($\rho = 4$, extrapolating one step past the range). Bars (left axis) show the share of the budget spent at each scale $C_j^\star / C$; the line (right axis) shows the number of runs $M_j^\star$. Cheap low scales absorb many runs but little compute; the budget concentrates on the high-leverage, low-$\gamma$ scale nearest the target ($u_4$), which is bought with only a few expensive runs. The near-central scale $u_2$ has $h_2 \approx 0$ and is kept only to anchor the fit.
    }
    \label{fig:compute-allocation}
\end{figure}

\paragraph{A manual for designing a compute-optimal scaling law estimate.}

Any practitioner should roughly follow the following steps in designing their experiment:

\begin{enumerate}
    \item  \emph{Estimate (or set) $\rho$}, the relative cost of a scale $N_j = N_1\,\rho^{\,j-1}$. For transformers this cost is well approximated by the standard $6ND$ estimate (roughly six FLOPs per parameter per token over the forward and backward pass) \citep{Kaplan2020ScalingLaws, Hoffmann2022AnEmpirical}, so once a practitioner chooses a model ladder and its token counts, every $N_j$, and hence $\rho$, is fixed. (Anyone building their own ladder has to do this FLOP accounting up front to know what a run at each scale will cost.) 
    \item \emph{Assume the effective dimensions $\gamma_j$.} We take these as known. This is not a perfect assumption, but it is broadly understood where along a model ladder tuning becomes reliable and $\gamma$ drops, and the choice can be checked after the fact from the realized slope of error against compute at each scale. 
    \item \emph{Compute the total budget $C$.} This is normally known outright usually in GPU-hours, converted to FLOPs through the hardware's sustained throughput. The only remaining factor, the per-scale error amplitude $A_j$, varies slowly and is typically taken comparable across scales (or read off the loss spread if not).
\end{enumerate}

With $\rho$, the $\gamma_j$, and $C$ in hand, the number of runs per scale can be computed. The multiplier $\lambda$ is computed through \Cref{eq:opt-runs}, and it is set by substituting the $M_j^\star$ into the budget constraint $\sum_j M_j N_j = C$ (\Cref{eq:constraint}) and solving the resulting equation for $\lambda$. Reading the $M_j^\star$ back out of \Cref{eq:opt-runs} then gives the number of tuning runs to launch at each scale.

%% file: contents/appendix/additional_empirical_results.tex
\section{Additional Empirical Results}
\label{sec:additional-results}

\begin{figure}[h]
    \centering
    \includegraphics[width=0.68\linewidth]{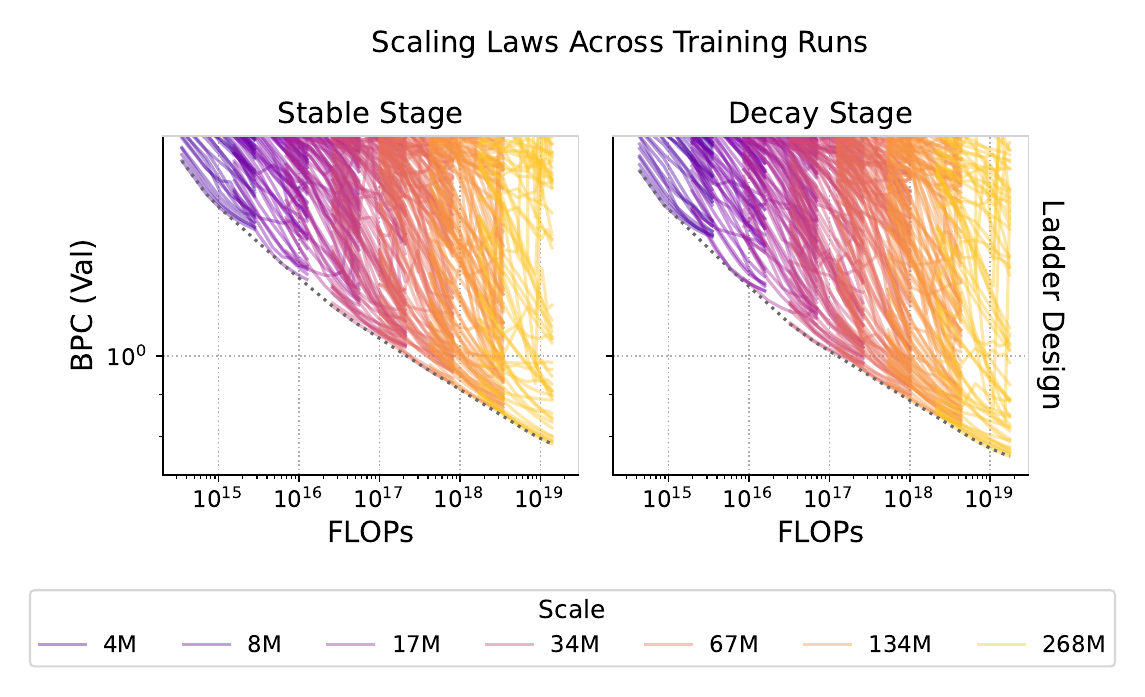}
    \caption{
        \emph{Scaling laws emerge at very small scales, smaller than widely believed; however, hyperparameter tuning becomes critical.} Predictable scaling appears only at the fully tuned frontier. Smaller models are harder to tune, making their scaling laws harder to observe though no less present.
    }
    \label{fig:scaling-laws-at-the-fully-tuned-frontier}
\end{figure}

\begin{figure}[h]
\vspace{-1cm}
    \centering
    \includegraphics[width=0.7\linewidth]{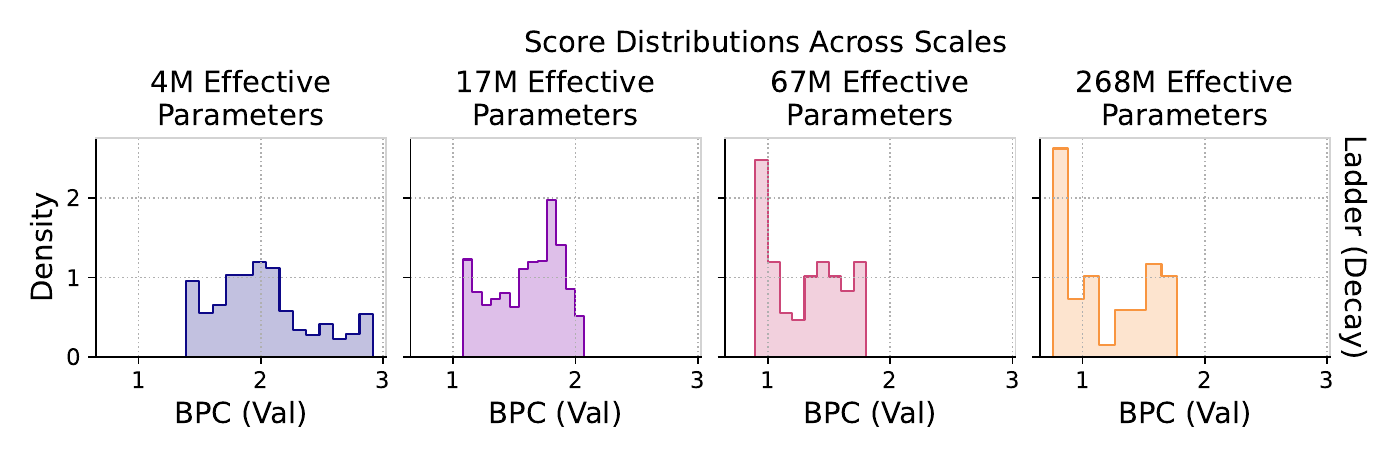}
    \caption{
        \emph{As scale increases, random configurations are more likely to land near the optimum.} We measure the volume of near-optimal configurations by visualizing the distribution over validation scores at the end of training from random search (the best 85\% to avoid divergent runs). As scale grows, the distribution concentrates near the best achievable loss at each scale, with probability mass piling up into a peak. Good configurations fill the space and get easier to find.
    }
    \label{fig:hyperparameter-sensitivity-decreases-with-scale}
\end{figure}

\begin{figure}[h]
\vspace{-1cm}
    \centering
    \includegraphics[width=\linewidth]{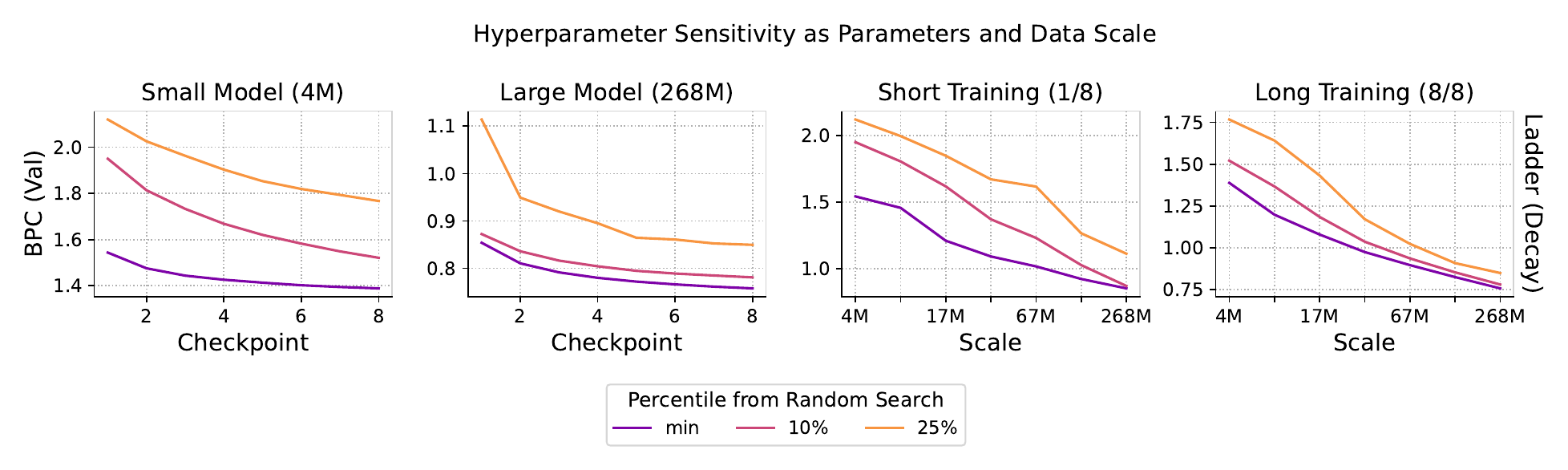}
    \caption{
        \emph{Hyperparameter sensitivity diminishes through increasing parameters and data together.} We plot the minimum, 10th, and 25th percentile validation losses from random search. The gap from the minimum measures sensitivity. Scaling data with a small model (left) or parameters with short training (middle right) lowers the loss but only slightly cuts the gap. Scaling data with a large model (middle left) or parameters with longer training (right) collapses the gap.
    }
    \label{fig:hyperparameter-sensitivity-decreases-through-both-parameters-and-data}
\end{figure}

% \begin{figure}
%     \centering
%     \begin{minipage}{0.48\textwidth}
%         \centering
%         \includegraphics[width=\textwidth]{figures/effective-hyperparameters-decrease-with-scale.pdf}
%         \caption{
%             \emph{Big models have fewer effective hyperparameters.} The effective hyperparameter count ($\gamma$) is the intrinsic dimension of the hyperparameter loss surface around the optimum. We estimate $\gamma$ by fitting the noisy quadratic to random search results. $\gamma$ drops to 1 as models scale.
%         }
%         \label{fig:effective-hyperparameters-decrease-with-scale}
%     \end{minipage}
%     \hfill
%     \begin{minipage}{0.51\textwidth}
%         \centering
%         \includegraphics[width=\textwidth]{figures/effective-hyperparameters-by-parameters-and-tokens.pdf}
%         \caption{
%             \emph{Parameters, more than data, reduce the effective number of hyperparameters.} The effective hyperparameter count ($\gamma$) is the intrinsic dimension of the hyperparameter loss surface. We estimate $\gamma$ at each parameter--data budget by fitting the noisy quadratic distribution to results from random search. With and without learning rate decay, $\gamma$ trends lower as parameters increase, while increasing data has much less of an effect.
%         }
%         \label{fig:effective-hyperparameters-by-parameters-and-tokens}
%     \end{minipage}
% \end{figure}

%% file: contents/appendix/case_study.tex
\section{Case Study Diagnostics}

\begin{figure}[!hbt]
    \centering
    \includegraphics[width=\linewidth]{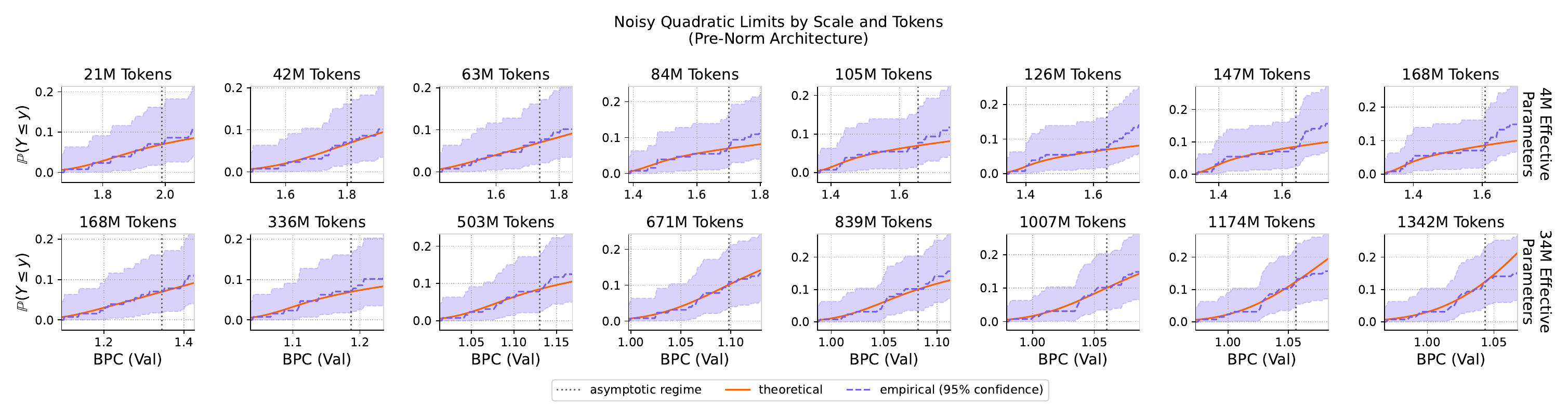}
    \includegraphics[width=\linewidth]{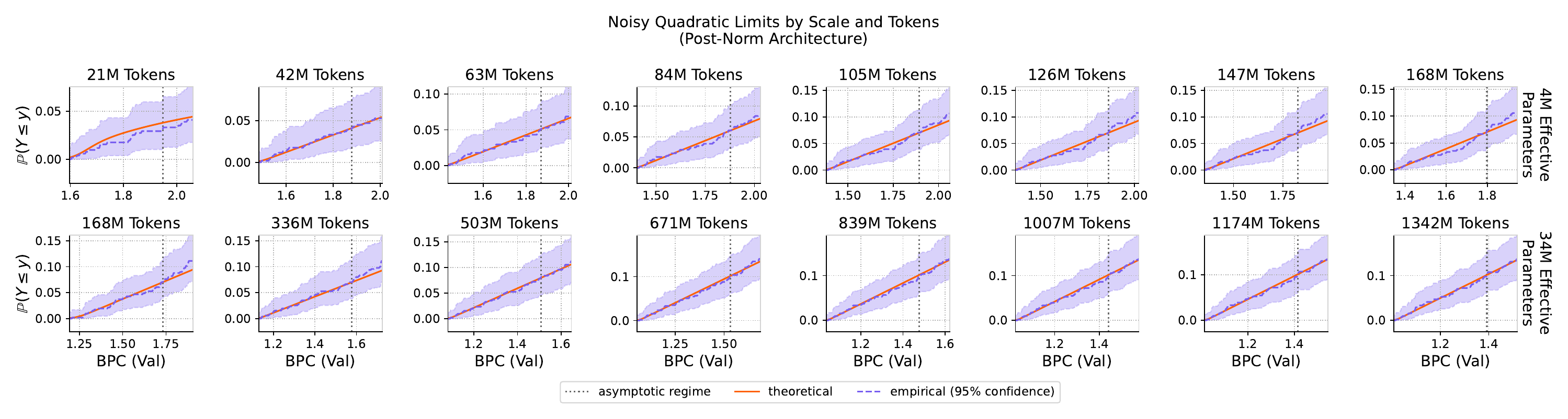}
    \caption{
        \emph{The noisy quadratic limit emerges for both architectures, confirming the hyperparameters are well-tuned.} We fit the noisy quadratic to the distribution of validation losses from random search at 4M and 34M parameters for the pre-norm (top) and post-norm (bottom) architectures. The asymptotic regime is smaller for post-norm than pre-norm, an early sign of its hyperparameter sensitivity.
    }
    \label{fig:noisy-quadratic-limit-for-prenorm-vs-postnorm}
\end{figure}
\begin{figure}[!hbt]
    \centering
    \begin{minipage}{0.49\linewidth}
        \centering
        \includegraphics[width=\linewidth]{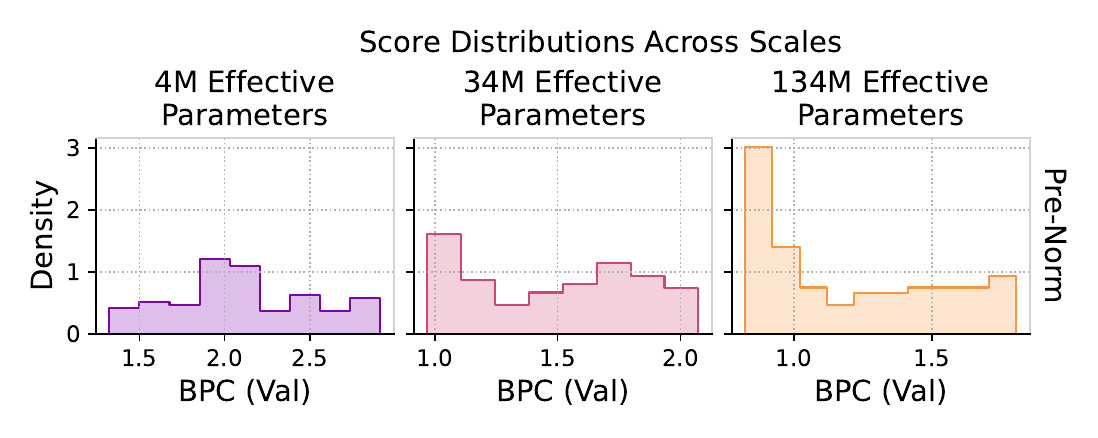}
    \end{minipage}
    \hfill
    \begin{minipage}{0.49\linewidth}
        \centering
        \includegraphics[width=\linewidth]{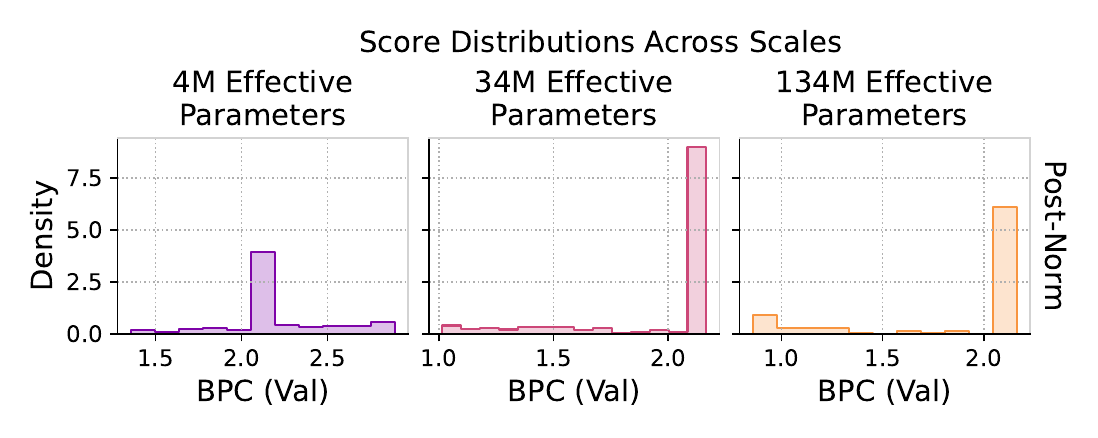}
    \end{minipage}
    \caption{
        \emph{Hyperparameter sensitivity decreases with scale for both the pre-norm and post-norm architectures.} Across scales, we compare the score distributions (best 85\%, to exclude divergent runs) for the pre-norm (left) and post-norm (right) architectures. Both place more probability mass near the optimum as scale increases; post-norm, however, maintains a large peak at a suboptimal value, showing persistent tuning difficulty.
    }
    \label{fig:sensitivity-across-scales-for-prenorm-vs-postnorm}
\end{figure}
\begin{figure}[!hbt]
    \centering
    \begin{minipage}{0.49\linewidth}
        \centering
        \includegraphics[width=\linewidth]{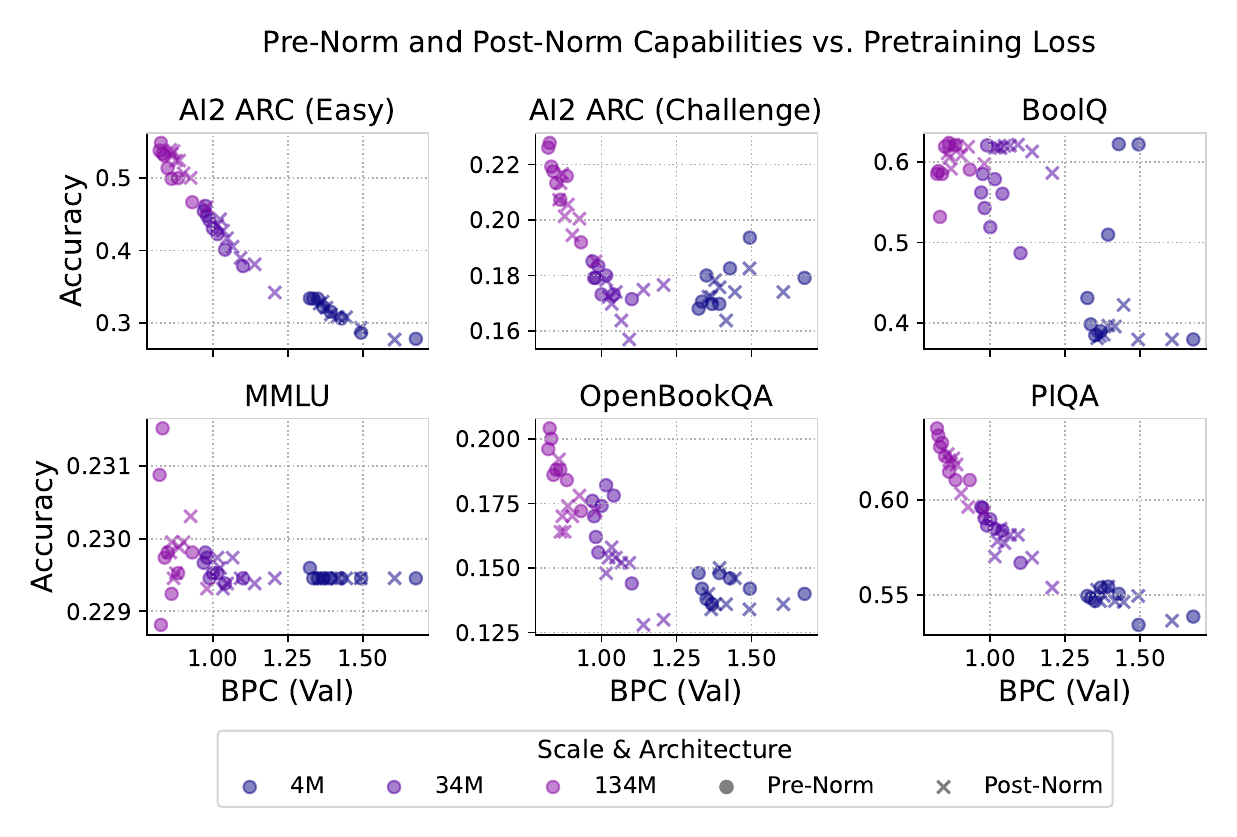}
        \caption{
            \emph{Pretraining loss corresponds to capabilities both within and across pre-norm and post-norm architectures.} Each panel plots pretraining loss against accuracy for a different task. Across scales and architectures, all points fall on a common trend: equal loss implies equal capability.
        }
        \label{fig:perplexity-capability-correspondence-for-prenorm-vs-postnorm}
    \end{minipage}
    \hfill
    \begin{minipage}{0.49\linewidth}
        \centering
        \includegraphics[width=\linewidth]{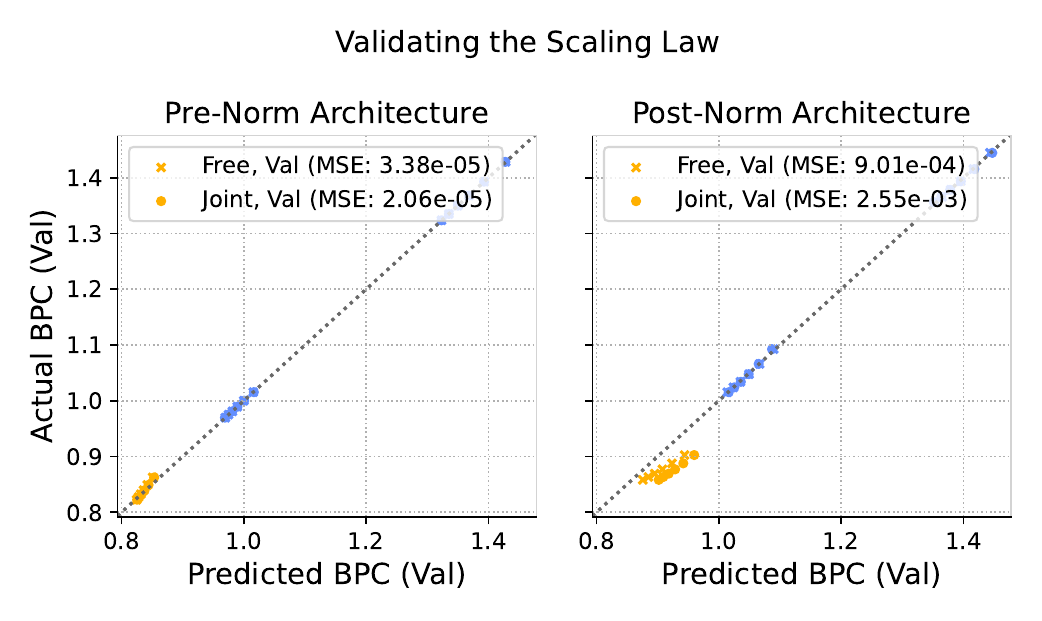}
        \caption{
            \emph{The scaling laws extrapolate comparably well with free vs. joint irreducible error terms.} We compare sharing the irreducible error term across the architectures (\texttt{joint}) vs. allowing them to differ (\texttt{free}). The pre-norm scaling law extrapolates very well in both cases. The post-norm law extrapolates somewhat, but could likely benefit from even further hyperparameter tuning.
        }
        \label{fig:validating-the-scaling-law}
    \end{minipage}
\end{figure}
\begin{figure}[!hbt]
    \centering
    \includegraphics[width=0.6\linewidth]{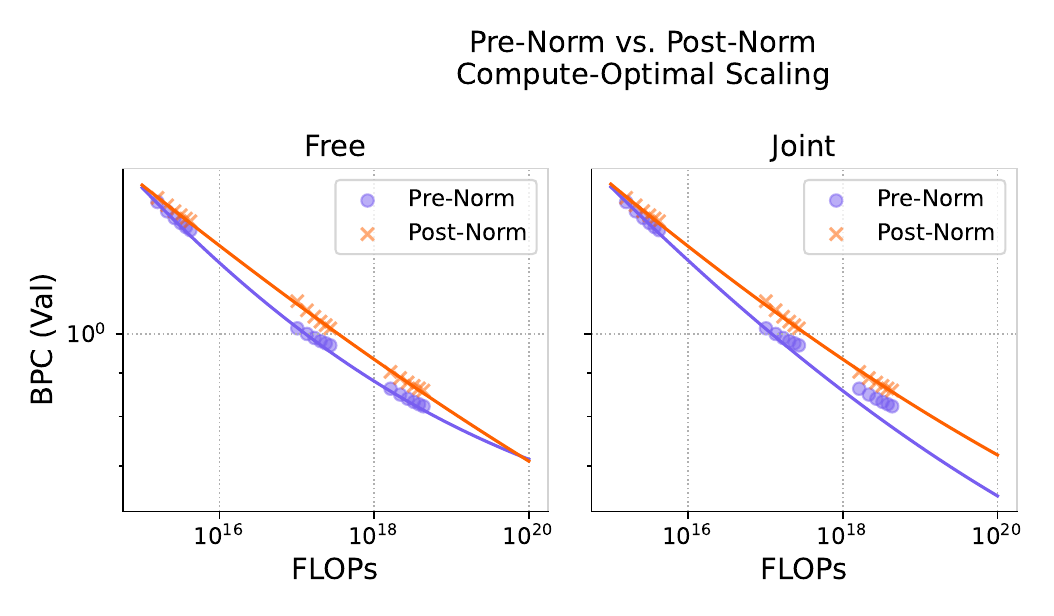}
    \caption{
        \emph{Pre-norm scales better than post-norm near the data; far from it, extrapolation reflects the irreducible error.} We fit the scaling laws to all available data and compare them under two conditions: separate (left) vs. shared (right) irreducible errors. Across the range we explore, pre-norm scales better. Outside that range, pre-norm continues to scale better if you believe that both architectures saturate at the same final loss. Under this assumption, the gap grows with compute and we recover the large-scale advantage of pre-norm from our small-scale experiments.
    }
    \label{fig:prenorm-vs-postnorm-compute-optimal-scaling}
\end{figure}